\documentclass[lettersize,journal]{IEEEtran}
\usepackage{soul}
\usepackage{marginnote} 

\renewcommand*{\marginnote}[1]{} 
\usepackage[dvipsnames]{xcolor}
\definecolor{pink}{HTML}{ec008c}

\usepackage[color=purple!15]{todonotes}

 \newcommand{\rev}[1]{{\color{black} #1}}

\newcounter{fsubsub}

\newcommand{\fsubsubsection}[1]{%
  \stepcounter{fsubsub}
  \subsubsection*{\textbf{(F\thefsubsub)} #1}
  \addcontentsline{toc}{subsubsection}{F\thefsubsub) #1}
}
\usepackage{amsmath,amsfonts}
\usepackage{algorithmic}
\usepackage{array}
\usepackage[caption=false,font=normalsize,labelfont=sf,textfont=sf]{subfig}
\usepackage{textcomp}
\usepackage{stfloats}
\usepackage{url}
\usepackage{verbatim}
\usepackage{graphicx}
\def\BibTeX{{\rm B\kern-.05em{\sc i\kern-.025em b}\kern-.08em
    T\kern-.1667em\lower.7ex\hbox{E}\kern-.125emX}}
\usepackage{balance}
\usepackage{enumitem}

\usepackage{booktabs}

\usepackage{xcolor}
\usepackage{tcolorbox}
\definecolor{semantic}{HTML}{955035}
\definecolor{primitive}{HTML}{024A70}
\usepackage{tabularx}
\usepackage{longtable}
\usepackage{listings}
\tcbuselibrary{breakable}
\usepackage{multicol}
\usepackage{wrapfig}

\usepackage{hyperref}
\definecolor{lightgray}{rgb}{0.92, 0.92, 0.92}
\newtcolorbox{promptbox}{
  colback=lightgray,
  boxrule=0pt,
  arc=0mm,
  fontupper=\ttfamily\scriptsize,
  left=1mm,
  right=1mm,
  top=1mm,
  bottom=1mm,
  breakable=true,
  parbox=false
}
\newcommand{\system}{VIDEE}

\makeatletter
\def\input@path{{sections/}}
\graphicspath{{figs/}{pictures/}{images/}{./}} 
\begin{document}
\title{VIDEE: Visual and Interactive Decomposition, Execution, and Evaluation of Text Analytics with Intelligent Agents}
\author{Sam Yu-Te Lee,
Chenyang Ji,
Shicheng Wen,
Lifu Huang,
Dongyu Liu,
Kwan-Liu Ma
\thanks{
Sam Yu-Te Lee, Shicheng Wen, Lifu Huang, Dongyu Liu, and Kwan-Liu Ma are with the Department of Computer
Science, University of California at Davis. E-mail: {ytlee, scwen, lfuhuang, dyuliu, klma}@ucdavis.edu; 
Chenyang Ji is an independent researcher.
Email: chenyangjizju@gmail.com
}
}

\maketitle

\begin{abstract}
Text analytics has traditionally required specialized knowledge in Natural Language Processing (NLP) or text analysis, presenting a technical barrier for analysts. 
Recently, agentic systems have shown great success in many automation tasks by significantly lowering the technical barriers, \rev{yet human insight is still needed.} 
In this work, we introduce VIDEE, a system that supports data analysts to visually control agents in text analytics. VIDEE instantiates a human-agent collaboration workflow with three stages: (1) Decomposition, which incorporates a human-in-the-loop Monte-Carlo Tree Search algorithm to support generative reasoning with human feedback, (2) Execution, which generates an executable text analytics pipeline, and (3) Evaluation, which integrates LLM-based evaluation and visualizations to support validation of execution results. 
We conduct two quantitative experiments to evaluate VIDEE's effectiveness and analyze common agent errors. A user study involving participants with varying levels of NLP and text analytics experience---from none to expert---demonstrates the system's usability and reveals distinct user behavior patterns. Our findings identify design implications for human-agent collaboration, and inform future improvements to intelligent text analytics systems.
\end{abstract}


\section{Introduction}
To extract useful information and insights from a large collection of textual data, 
the ability to choose and conduct a sequence of text analytics tasks, from 
named entity recognition, sentiment analysis, to document clustering~\cite{Liu2019textminingsurvey, talib2016textmining},  
is essential.  
In the past, to carry out these tasks, one must acquire advanced Natural Language Processing (NLP) knowledge for a comprehensive understanding of the common text analytics methods, creating significant barriers for data analysts.

\rev{
\marginnote{$\triangle$\_1\_1}
\rev{Recent advances in prompt engineering and agentic systems} have brought significant changes to text analytics. 
Many conventional text analysis methods can now be implemented with prompt templates~\cite{xu2024llmforie, Wan2024TnTLLM}, and coding agents~\cite{anthropic2024claudecode} promise to iteratively generate and execute codes to perform text analytics autonomously. 
However, our preliminary experiment on coding agents reveals several concerning issues regarding fully autonomous text analytics agents, such as over-simplification, hallucination, and a lack of robust architecture, suggesting the necessity for human oversight and intervention~\cite{passi2025agentic_ai_human_oversight}.
Still, enabling human oversight and intervention in the agentic text analytics introduces new challenges.} 

\textit{\textbf{Large decomposition space:}}
Text analytics allows for many ways of decomposing a goal into subtasks.
Analysts need to balance the trade off between the difficulty of subtasks and the overall robustness of the pipeline, as excessively long and fragmented pipelines may increase computational cost and \mbox{risk of errors}.
\rev{However, agents tend to oversimplify the goal and regard an ill-considered plan as definitive, which can easily mislead inexperienced analysts.}

\textit{\textbf{Varying technical knowledge:}}
LLM-related fields are undergoing rapid advancements~\cite{promptengineeringuide}.
Analysts, possess varying levels of technical knowledge, might not keep up with the state-of-the-art knowledge, such as Retrieval-Augmented-Generation (RAG)~\cite{lewis2020rag}, generative reasoning~\cite{xie2023beamsearch}, context engineering~\cite{mei2025surveycontextengineering}, making new advancements under-utilized.

\rev{
\textit{\textbf{Constructing a robust pipeline:}} 
While agents perform well on individual tasks, they are still unreliable in connecting multiple tasks into a text analytics pipeline, let alone experimenting with alternative pipelines.
The analyst needs to have access to the input and output formats, intermediate data transformations, and parameters in each step. 
}

In response, we propose and investigate the effectiveness of a human-agent collaboration workflow with three stages: \textit{Decomposition, Execution}, and \textit{Evaluation}, instantiated by \system \ \footnote{\url{https://github.com/SamLee-dedeboy/VIDEE}}.
Given a user goal and a dataset, the agent, controlled by a human, \textbf{decompose} the goal by searching the large decomposition space and generate a text analytics plan. 
The agent then \textbf{implement} individual tasks as planned and \textbf{execute} them upon request, while the human uses a set of criteria to \textbf{evaluate} the execution performance and refine the pipeline.
In the end, the human derives insights from the pipeline results.

We conducted quantitative experiments to evaluate the robustness and reliability of the system, and a qualitative user study with participants ranging from no experience to expert-level experience with NLP or text analytics for usability and utility evaluation.
\rev{We report the user behavior patterns that provide insights in the human-agent collaboration workflow, and discuss implications on how visualizations can assist in supporting human oversight and control.}
Our contributions are:
\begin{itemize}
    \item A three-stage human-agent collaboration workflow for text analytics;
    \item \system, an agentic system with a visualization interface that supports data analysts perform text analytics, and 
    \item Empirical findings from system evaluations and insights on the role of visualization in agentic systems.
\end{itemize}



\section{Related Work}

\subsection{Text Analytics with LLMs}
Researchers have found that LLMs can achieve state-of-the-art performance in many individual text analysis tasks, from classification and annotation~\cite{fabrizio2023chatgptannotation, törnberg2023chatgpt4outperformsexpertscrowd} to universal information extraction~\cite{xu2024llmforie} and knowledge graph construction~\cite{zhang2024llmkgc}.
The findings of these works coincide with other large-scale evaluations of LLMs in NLP~\cite{chang2024surveyllm}, suggesting that LLMs are ready to be text analysis task solvers in real-world settings. 

Beyond individual analysis, LLMs have been increasingly used to construct end-to-end text analytics pipelines.
Wan et al.~\cite{Wan2024TnTLLM} proposed TnT-LLM, a framework for LLM-based label generation and annotation for automatic text mining. Lam et al.~\cite{lam2024lloom} developed LLooM, a system that  summarizes high-level concepts from unstructured text by chaining multiple prompt-based ``operators'' into a pipeline. 
Due to its strong flexibility, LLM-based text analytics pipelines can generalize beyond conventional NLP tasks, such as qualitative thematic analysis~\cite{dai2023llmintheloop, rasheed2024llmforQDA}.
While promising, constructing such pipelines still require significant effort. In this work, we support the semi-automatic generation of such pipelines under a human-agent collaboration paradigm.



\rev{\subsection{Agentic Data Analysis and Visualization}}
Researchers have found LLMs useful in assisting or even automating data analysis processes, e.g., the Data Wrangler~\cite{microsoftDataWrangler}, which supports data cleaning and transformation with natural language.
\rev{
More recently, coding agents~\cite{anthropic2024claudecode} that are capable of generating and executing codes iteratively shows strong potential in automating data analysis tasks.  
}
However, Kazemitabaar et al.~\cite{kazemitabaar2024llmforda} found that users face serious challenges in verifying AI-generated results and steering AI in data analysis, calling for greater human involvement.

\rev{
\marginnote{$\triangle$\_2\_1}
The method of leveraging autonomous agents to iteratively generate and interpret visualizations for data analysis is increasingly referred to as agentic visualization~\cite{dhanoa2025agentic_visualization}.
While study shows that visualization helps agents understand data~\cite{Li2025vis_help_ai_understand_data}, agentic visualization systems still suffers from several limitations, such as inconsistent visualization designs~\cite{maddigan2023Chat2Vis}, and unfaithful or factually incorrect interpretations~\cite{hong2025LLM_VL_modified_VLAT, dong2025VL_VLM}.

}

\rev{
Compared to tabular data, text data analysis and visualization present unique challenges and our preliminary experiment shows that current coding agents are not yet reliable for text analytics. In response, we propose a human–agent collaborative workflow in which human retains control over task decomposition, execution, evaluation, and result interpretation.

}


\subsection{Human Oversight and Control of Agentic Systems}
\rev{
While agentic systems are expected to function independently of human, researchers have found that agents still make mistakes~\cite{cemri2025MAST} and human oversight remains necessary~\cite{passi2025agentic_ai_human_oversight}.
Thus, designs that enhance human control and evaluation in agentic interfaces are essential, yet remain under-explored.}
Cai et al.~\cite{cai2024lowcodellmgraphicaluser} introduce Low-code LLM, a framework integrating multiple types of visual programming to support users edit the agent's plan before execution. 
Xie et al.~\cite{xie2024waitgpt} transform LLM-generated data analysis code into an interactive visual representation to enhance user comprehension and control. 
Cheng et al.~\cite{Cheng2024relic} designed RELIC, an interactive system that supports evaluation of factuality in agent responses based on the consistency of the responses.
However, it remains technically challenging to reliably assess the uncertainty of agent responses in general~\cite{beigi2024uncertaintyreview}, and therefore LLM judges\cite{chiang-lee-2023-large, liu-etal-2023-g}, i.e., defining evaluation criteria as LLM instructions in prompt templates, are more widely adopted.
These works collectively demonstrate the need for human-centered interfaces in agentic systems. We build on these insights and design our system specifically for the unique requirements of text analytics.



\begin{table*}[htbp]
\centering
\begin{tabular}{c | c | m{5cm} | m{4cm} | c | c}
\toprule
\textbf{Session} & \textbf{Type} & \textbf{Seed Prompt} & \textbf{Research Question} & \textbf{Success?} &\textbf{Overall Score} (1-10) \\
\hline
1-1 & Individual Tasks & \textit{Extract all named entities from these news articles.}
& Can the agent implement NER? & Yes & 8.25 \\
\hline
1-2 & Individual Tasks & \textit{Who are the most mentioned people in these news articles?}
& Can the agent formulate an NER task from non-technical terms? & Yes &8.8\\
\hline
1-3 & Individual Tasks  & \textit{What topics dominate the news coverage?} 
& Can the agent implement topic modeling? & No &7.6\\
\hline
1-4 & Individual Tasks & \textit{How do these news sources differ in their coverage priorities?}
& Open-ended task with ambiguity. & No &3.8\\
\hline
2-a-[1,2,3] &Pipeline &  \textit{For each major entity mentioned in the articles, analyze the sentiment of sentences discussing that entity, and report statistics.}
& Can the agent consistently compose a pipeline when requested explicitly? & No & [7.7, 6.7, 5.5]\\
\hline
2-b-[1,2,3] & Pipeline &  \textit{Analyze media bias in these news articles.}
&  Can the agent recognize ambiguity in the request and compose a pipeline? 
& No & [4.3, 3.1, 6.3] \\
\bottomrule
\end{tabular}
\vspace{0.1cm}
\caption{Overview of experimental sessions and their corresponding seed prompts and expectations.}
\label{tab:sessions}
\end{table*}

\rev{
\section{Preliminary Experiment: Coding Agents}
\marginnote{$\triangle$\_3\_1}
In this section, we report the findings of an experiment on how LLM-based coding agents perform in text analytics. We found several severe issues, such as hallucinating results and problematic analytical decisions, suggesting that fully autonomous agents are not yet ready to conduct text analytics end-to-end. 
\subsection{Experiment Procedure}
We conducted 10 sessions with Claude Code~\cite{anthropic2024claudecode} using ``Claude 3.7 sonnet'', testing two types of tasks: text analysis with individual tasks (4 requests with increasing difficulty) and with a pipeline (2 requests repeated 3 times), as shown in ~\autoref{tab:sessions}.
We test two requests on the pipeline tasks (entity-sentiment analysis and media bias analysis) and repeat each request three times to evaluate consistency. 
In each session, the agent is given a prompt consisting of the dataset location, the seed prompt, and a request to save all intermediate results. 
The agent is expected to generate and execute code to read the data, conduct analysis, document intermediate results, interpret results, and write a human-readable report. 

The dataset consists of 2000 news articles sampled from the AllTheNews dataset~\cite{thompson_allthenews}.
For each session, we manually produce an evaluation report based on the conversation history and all files generated by the agents.
The report rates each session from 1 to 10 with a set of criteria, e.g., methodological soundness, correctness of the generated codes, validity of interpretation, and readability of reports. 
These scores are meant for keeping track of performance variations rather than robust quantitative evaluation. Our findings are primarily derived from key observations. All agent responses and evaluation reports are in the supplemental material. 

\subsection{Main Findings}
In general, the agents can perform individual tasks with a relatively high success rate and quality. 
However, the quality degrades significantly when the seed prompt is open-ended (session 1-4) and in pipeline tasks. 
We summarize the failure modes in the following paragraphs.

\fsubsubsection{\textbf{Over-simplification of user request}}
Despite the seed prompt being open-ended, e.g., topic modeling can be accomplished using a variety of methods, the agent would always present the proposed method and results as definitive, without acknowledging limitations or proposing alternatives. For pipeline tasks, the agent generally took a ``kitchen-sink'' approach, i.e., implementing multiple analysis methods without a coherent analytical framework, nor did it logically combine multiple analysis results for interpretation.
The agents tend to oversimplify user requests confidently.

\fsubsubsection{\textbf{Problematic analytical decisions and implementation}}
We found several instances of problematic analytical decisions. For example, to derive topics in session 1-4, the agent assumes certain topics exist and uses hardcoded topic keywords to collect statistics instead of using topic modeling techniques.
Regarding implementation, the agent sometimes misuses NLP tools. In session 1-4, the agent extracts keywords by concatenating 2000 articles into one text document and employs a vectorization method based on word frequency. Then, it disregards the vectorized representation and retrieves feature names and counts. This is equivalent to directly calculating word frequency, but is unnecessarily complicated and confusing. 

\fsubsubsection{\textbf{Hallucination of code execution}}
In 3 out of 6 pipeline sessions, the agent did not execute the generated code. Moreover, the agent would hallucinate by presenting analytical findings as if the code were executed. 
For example, in session 2-a-1, the agent provides a ``Status Report'' mentioning ``Anthony Fauci (156 mentions)'' and ``Kamala Harris'' (132 mentions), but both are fabricated data.
Despite a lack of evidence, the presented findings appear to be professional and easy-to-read, making it convincingly misleading. This finding raises serious concerns regarding the use of fully autonomous agents in analytical tasks.

\fsubsubsection{\textbf{Limited observability}}
Claude Code's conversational interface makes it hard to detect agent mistakes.
In session 2-b-a, the agent conducted exploratory analysis, but the intermediate results and findings are not stored, making the analytical process non-auditable and obscuring potential reasoning errors.
In session 2-a-2, the agent calculates entity frequency by first concatenating 2000 articles and truncating characters at 1 million characters before NER, resulting in silent data loss as only 500-700 articles are included.


\fsubsubsection{\textbf{Unreliability under context window saturation}}
When handling large datasets, the agent's behavior becomes unreliable. First, the agent struggles to manage long-running background tasks. While waiting for processes to complete, it attempts to sustain conversation activity by generating unnecessary auxiliary work, such as drafting demo scripts or summarizing intermediate progress. As these additional outputs accumulate and exceed the context window, the agent begins to exhibit unreliable behavior due to context loss.
In one instance, the agent unilaterally substituted the output of a demo script (run on only 50 articles) for the full analysis and prematurely declared the task complete. 

\fsubsubsection{\textbf{Coding errors, waste, and inconsistency}}
Despite operating under the ReAct~\cite{yao2023react} paradigm, the agent still generates codes with runtime errors that, while all fixed reactively, raise concern over code reliability and cause significant regeneration waste. For example, to modify entity types used to a filtering function, the agent generates a 475-line script from scratch instead of making a targeted modification. 
As we repeat session 2-a and 2-b, we found significant inconsistency as the agent chooses three fundamentally different analytical pipelines consisting of different tools and metrics for the same request. This inconsistency suggests that the analysis methods are ill-considered and unjustified.


\subsection{Conclusion of experiment}
The preliminary experiment results show that current coding agents are not yet reliable to conduct text analytics end-to-end.
Human oversight is needed to ensure well-considered decomposition (F1) and analytical decisions (F2). A robust architecture is needed to prevent hallucination (F3), unreliable agent behavior (F5), and to reduce coding errors, waste, and inconsistency due to regeneration (F6). Moreover, a typical conversational interface lacks the affordances to support effective human oversight (F4), and thus a bespoke interface is needed for effective human-agent collaboration.
}

\begin{figure*}[t]
    \centering
    \includegraphics[width=\textwidth]{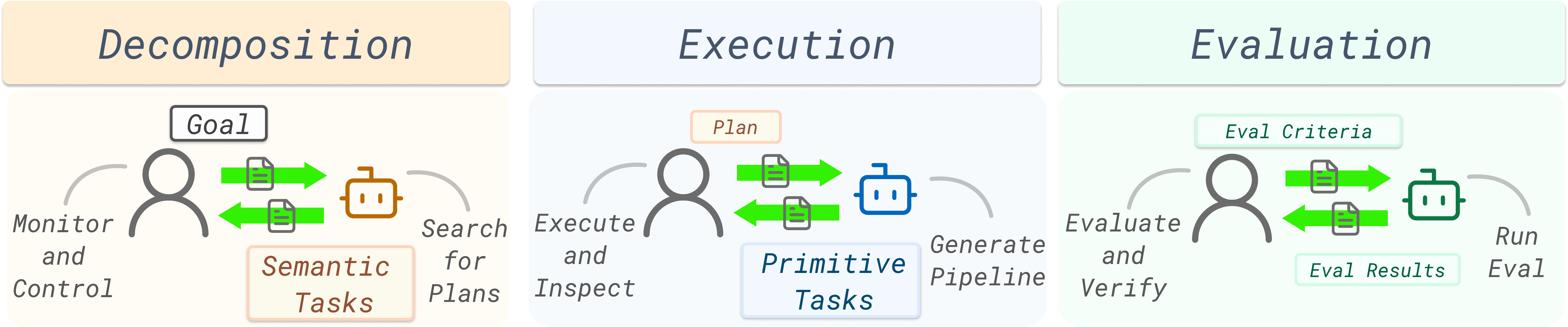}
    \caption{A three-stage human-agent collaboration workflow for text analytics with multiple agents. In the Decomposition stage, the human describes a goal and the decomposer agent searches for a plan under human monitor and control. The results are communicated back as semantic tasks (e.g., topic modeling). In the Execution stage, the executor agent generates a pipeline based on a plan specified in primitive tasks (e.g., cluster analysis or document classification) for the human to execute and inspect. In the Evaluation stage, the evaluator agent runs evaluation (e.g., topic coverage) using the human specified criteria and present the results for the human to evaluate and verify.
    }
    \label{fig: workflow}
    \vspace*{-0.5cm}
\end{figure*}

\rev{
\section{Design Analysis} 
We collaborate with senior NLP experts to design a human–agent collaboration workflow that integrates human judgment with agents’ coding and reasoning capabilities to support entry-level data analysts. The workflow defines specialized agents for text analytics, structures task planning, execution, and evaluation, and incorporates explicit interrupt points for human intervention. \system\ operationalizes this workflow through tailored visualization and interaction designs.
}

\subsection{Human-Agent Collaboration Workflow}
First, we define some terms involved in the workflow:
\begin{itemize}[leftmargin=4mm]
    \item \textbf{Goal}: A practical action to be achieved by analyzing the dataset, such as exploring the HCI research topics in recent years or understanding sentiments of user comments.
    \item \textbf{Semantic tasks}: Conceptual descriptions of how the goal is to be achieved through text analytics. This level of description is appropriate for high-level planning and communication with non-technical people.
    \item \textbf{Decomposition}: The process of searching for a plan consisting of semantic tasks in order to achieve the goal.
    \item \textbf{Primitive task}: A concrete definition of a text analysis method. It should contain implementation information such as the algorithm and the data schema. This level of description is appropriate for analysts to implement and debug.  
    \item \textbf{Pipeline}: A sequence of primitive tasks, including the definitions of each primitive task and their dependencies.
\end{itemize}
Next, we explain in detail the three-stage workflow of the system, as shown in~\autoref{fig: workflow}. For each stage, we highlight the division of responsibility between the human and the agent. 
Here, we focus on the high-level collaboration pattern, and leave the technical details in~\autoref{sec: system_architecture}.

\subsubsection{Decomposition} 
In this stage, the human and the agent work together to plan a pipeline consisting of \textit{semantic tasks} from the \textit{goal}. 
We refer to the agent at this stage as the \textbf{decomposer agent}. 
Given a user goal, the decomposer agent explores possible pipeline options in a generative reasoning approach~\cite{xie2023beamsearch} with 
a feedback-aware search algorithm governs the search direction.
The \textbf{human} user monitors the agent's search process, and controls the agent towards a desirable pipeline. 
At the end of decomposition, a sequence of \textit{semantic tasks} is finalized as the plan for the text analytics pipeline.


\subsubsection{Execution} 
In the execution stage, an \textbf{executor agent} generates an executable pipeline consisting of \textit{primitive tasks}.
Each primitive task specifies a text analysis method to conduct, necessary parameters, and the executables, i.e., a prompt template or an executable function.
The \textbf{human} verifies the pipeline, executes each step, and checks for undetected errors. 



\subsubsection{Evaluation}
In this step, a set of evaluation methods is performed on the execution result of each primitive task, and the evaluation results are presented with visualizations for human inspection.
To compensate for human neglect of evaluation~\cite{kim2024evallm}, the \textbf{evaluator agent} can recommend evaluation criteria for the \textbf{human} user.
In our system, each evaluation method is instantiated as an LLM judge that can be generated from a user-defined criterion expressed in natural language. 


\vspace*{-0.3cm}
\subsection{Scenario Walkthrough of \system} 
\begin{figure*}[t]
    \centering
    \includegraphics[width=\textwidth]{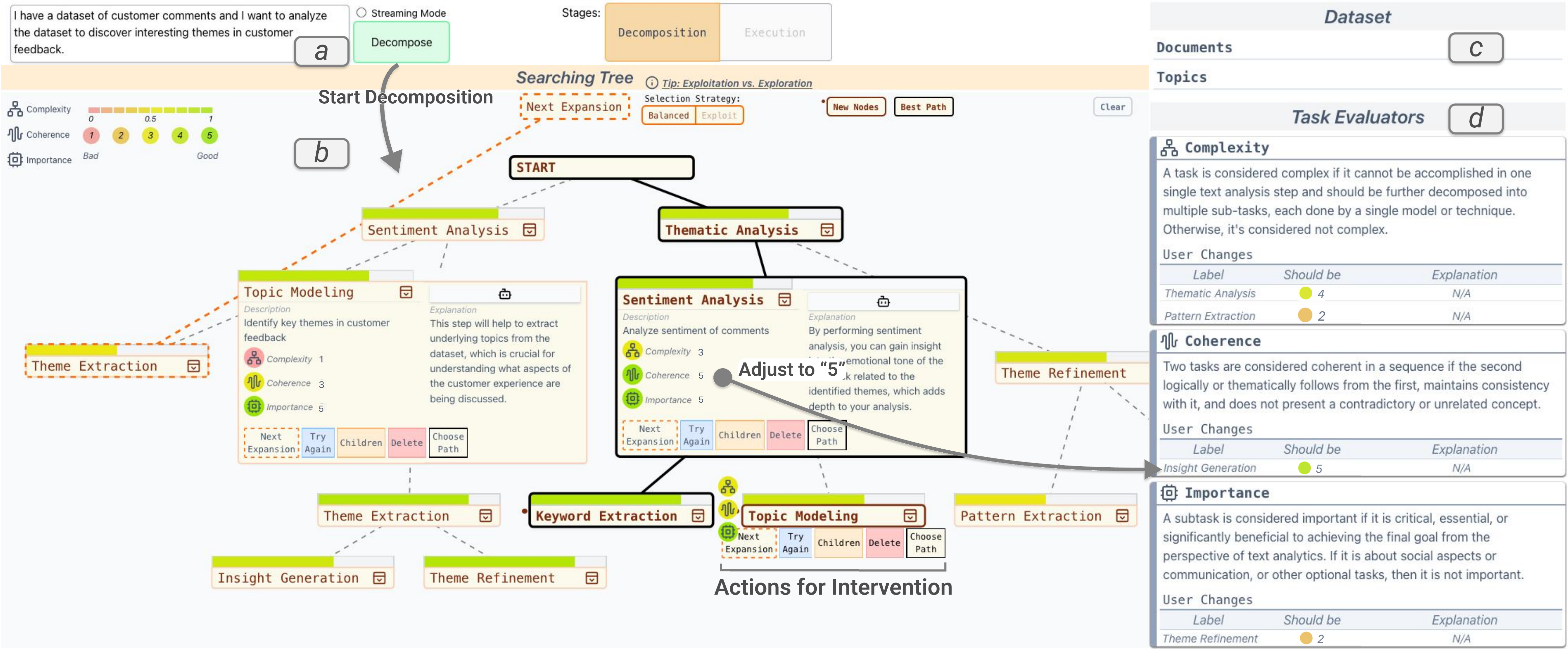}
    \caption{The interface for the decomposition stage. (a) Users can input their goal and dataset context in natural language. (b) The decomposer agent iteratively searches for text analytics plans to accomplish the goal using Monte-Carlo Tree Serch. Users can make various actions to intervene in the search process, such as choosing next expansion node or adjusting the scoring of a node. (c) Dataset Inspection View supports users to make sense of the dataset. (d) The scoring criteria of each node, including complexity, coherence, and importance. User feedback on the scoring is recorded by the system. }
    \label{fig: decomposition-overview}
    \vspace*{-0.5cm}
    
\end{figure*}

Based on the proposed workflow, we demonstrate how Alice, a junior data analyst, could use \system \ to accomplish the task of identifying key themes in customer feedback. 
As shown in~\autoref{fig: decomposition-overview}-a,
Alice starts by expressing her goal and dataset in natural language: ``\textit{I have a dataset of customer comments and I want to analyze the dataset to discover interesting themes in customer feedback}.'' 
This naturally leads Alice to the \textbf{Decomposition} stage, in which a decomposer agent searches for an analysis plan (\autoref{fig: decomposition-overview}-b).
As each node is defined as a semantic task, Alice can easily make sense of available options while monitoring the search progress.
Besides typical tasks such as sentiment analysis and topic modeling, Alice is exposed to a wide variety of alternatives, such as \textcolor{semantic}{Keyword Extraction} and \textcolor{semantic}{Pattern Extraction}. 
Alice can also control the agent in various ways, such as changing the search direction or adjusting how the agent evaluates the search results.

After some searching, Alice chooses a plan consisting of three semantic tasks: \textcolor{semantic}{Thematic Analysis}, \textcolor{semantic}{Sentiment Analysis}, and \textcolor{semantic}{Keyword Extraction}. 
Then, she moves on to the \textbf{Execution} stage (\autoref{fig: execution-evaluation-overview}-b).
Using the plan, the system automatically generates an executable pipeline with six steps, including \textcolor{primitive}{Embedding Generation}, \textcolor{primitive}{Clsutering Analysis}, \textcolor{primitive}{Label Generation}, and so on.
Alice simply needs to inspect and execute each node in the pipeline sequentially and make necessary changes to the execution parameters (\autoref{fig: execution-evaluation-overview}-d).
For example, \textcolor{primitive}{Sentiment Analysis} is implemented with a prompt template, and Alice can edit it to specify the customer attitudes she is interested in. 
After each execution, Alice shifts attention to \textbf{Evaluation} stage to assess the quality of the execution results. Alice can use the system recommended evaluators or add new ones, and inspect the results with visualizations (\autoref{fig: evaluation_results}).

After executing the whole pipeline, Alice discovers some interesting themes and sentiments (\autoref{fig: execution-evaluation-overview}-e).
For example, customers report positively on ``the Speed, Capacity, Performance, and Compatibility of microSD Cards'', and report negatively on ``Card Issues and Customer Support Experience''.
By conducting text analytics in the interface, Alice can accomplish her goal in a much more efficient and reliable way.

\section{Interface Design}
The visualization interface of \system \ instantiates the three-stage workflow, supporting a no-code environment to plan, execute, and evaluate a text analytics pipeline. 
The interface consists of two separate pages. The first page contains \textbf{Searching View} and \textbf{Dataset Inspection View}, supporting the \textit{Decomposition} stage of the workflow.
The second page consists of four views: \textbf{Plan View}, \textbf{Execution View}, \textbf{Evaluation View}, and \textbf{Inspection View}. The four views coordinate to support the \textit{Execution} and {Evaluation} stage.  
Next, we introduce each view in more detail and discuss their design considerations.

\vspace*{-0.3cm}
\subsection{Searching View} 
Searching View streams the MCTS process in a tree diagram and supports various user controls (\autoref{fig: decomposition-overview}-b). 
Starting with a goal input box positioned at the top(\autoref{fig: decomposition-overview}-a), semantic tasks are incrementally added to the canvas as the search proceeds, with each node representing one task (\autoref{fig: decomposition-overview}-b). 
Each node shows (1) a label and description, (2) an explanation for why this task is needed, and (3) scores for three criteria: \textit{ Complexity of task, Coherence with the parent node, and Importance for achieving the overall goal}. Each criterion is scored on a discrete Likert scale from 1 to 5 (higher being more desirable). 
These scores are used by the MCTS algorithm as reward values to guide the selection of expansion nodes.
More technical details are introduced in~\autoref{sec: system_decomposer}.

\subsubsection{Layout}
We employ the Sugiyama graph layout method~\cite{kazuo1995sugiyama} to visualize the search tree. Each expansion adds $k$ candidate semantic tasks, shown as child nodes. In this layout, any path from the root to a leaf represents a complete analytics plan, which users can select to proceed to the Execution stage. The layout avoids node overlaps and supports zooming and panning for easier exploration.

\subsubsection{Visual Encodings}
To help users monitor and steer the search process, we provide both textual and visual encodings. The three discrete scores (complexity, coherence, importance) are each represented by a unique icon.
The scores of each criterion is aggregated and normalized between $0$ and $1$, encoded as the horizontral bar at the top of each node, providing an overview.
Individual and aggregated scores share a red-to-green color scale with green indicating more desirable values to ensure visual consistency. 

We also distinguish three types of nodes in the visual design: (1) the node selected for expansion, (2) nodes generated from last iteration, and (3) nodes on the highest-value path. 
These encodings support users assess node quality, identify low-scoring branches, and control the search direction.

\subsubsection{User Interactions}
\begin{figure*}[!t]
    \centering
    \includegraphics[width=\textwidth]{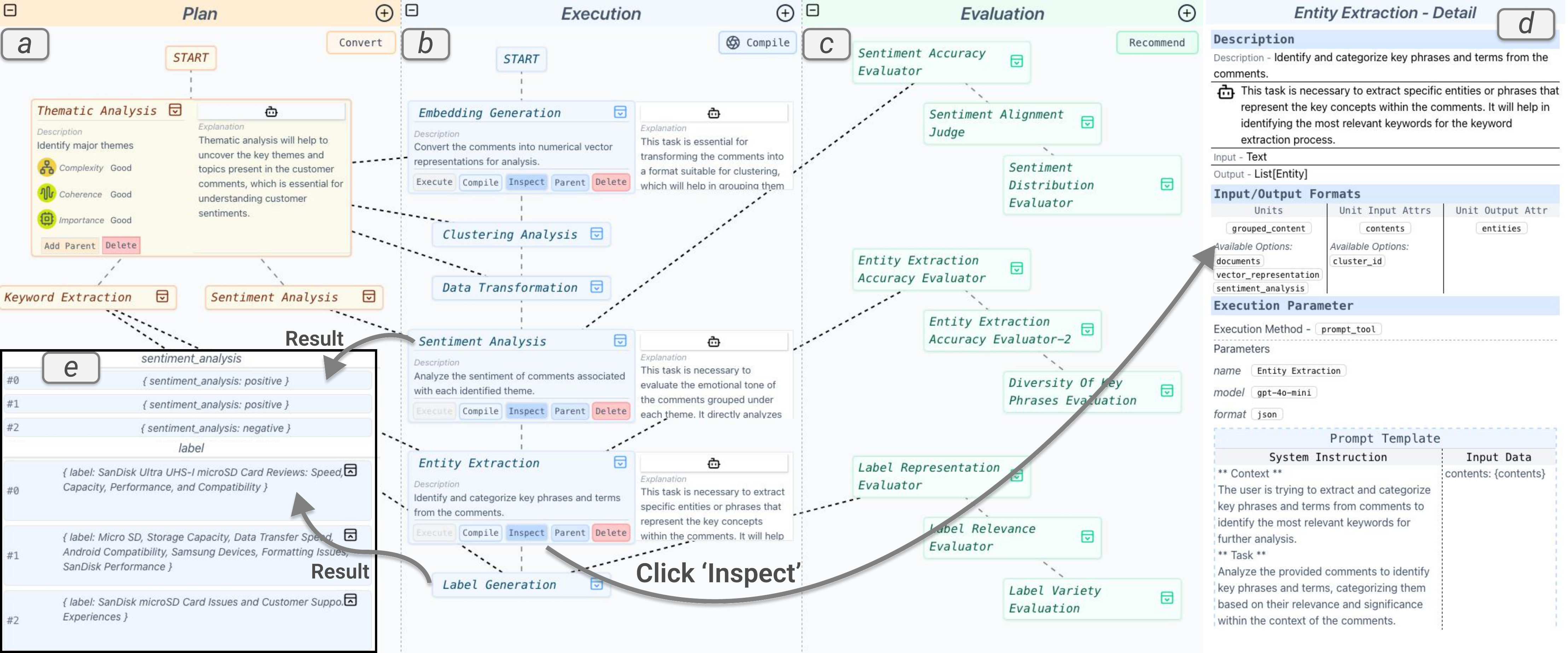}
    \caption{The interface for the execution and evaluation stage. (a) The user selected plan in the decomposition stage. Users can make final adjustments here, then click ``Convert''. (b) Based on the plan, the system generates an executable pipeline. Each node is a primitive task with label, description, and execution parameters. Users can click the ``Execute'' button to execute a node.
    (c) For each primitive task that needs evaluation, the system automatically recommends three evaluation criteria using LLM judges. Users can also add their own evaluation criterion using the ``+'' button on the top right corner. Each evaluation node is also executable. (d) The inspection panel showing the detail of a selected node. Users can see more detail of a node or make necessary changes, such as changing the input/output, execution parameters, or prompt templates. (e) Users can inspect the execution results in the interface.
    }
    \label{fig: execution-evaluation-overview}
    \vspace*{-0.5cm}
\end{figure*}

Searching View supports several kinds of user interaction.
(1) To \textbf{control} searching, users switch between streaming or step-by-step generation with the ``Streaming Mode'' button. They can also change the search strategy between ``Balanced'', which uses the UCT selection strategy~\cite{kocsis2006montecarlo2}, or ``Exploit'', which uses the greedy strategy.
(2) To \textbf{make sense of} the search results, users can expand a node to inspect its description, or hover over a score icon to inspect an AI-generated explanation.
(3) To \textbf{edit} the tree, users can regenerate or delete a branch, and add children to a node.
(4) To \textbf{provide feedback} on the reward, users can change the scores by clicking the criterion icon and entering an optional explanation, or directly editing the definition of a criterion (\autoref{fig: execution-evaluation-overview}-d). 
These changes are recorded and taken into account in future scoring.
(5) Users can \textbf{select a path} as the final plan, and move on to the execution stage by clicking the ``Execution'' button.

\vspace*{-0.2cm}
\subsection{Dataset Inspection View and Topic Radial Chart}\label{sec: dataset_inspection_view}
\begin{figure}[!t]
    \centering
    \includegraphics[width=\columnwidth]{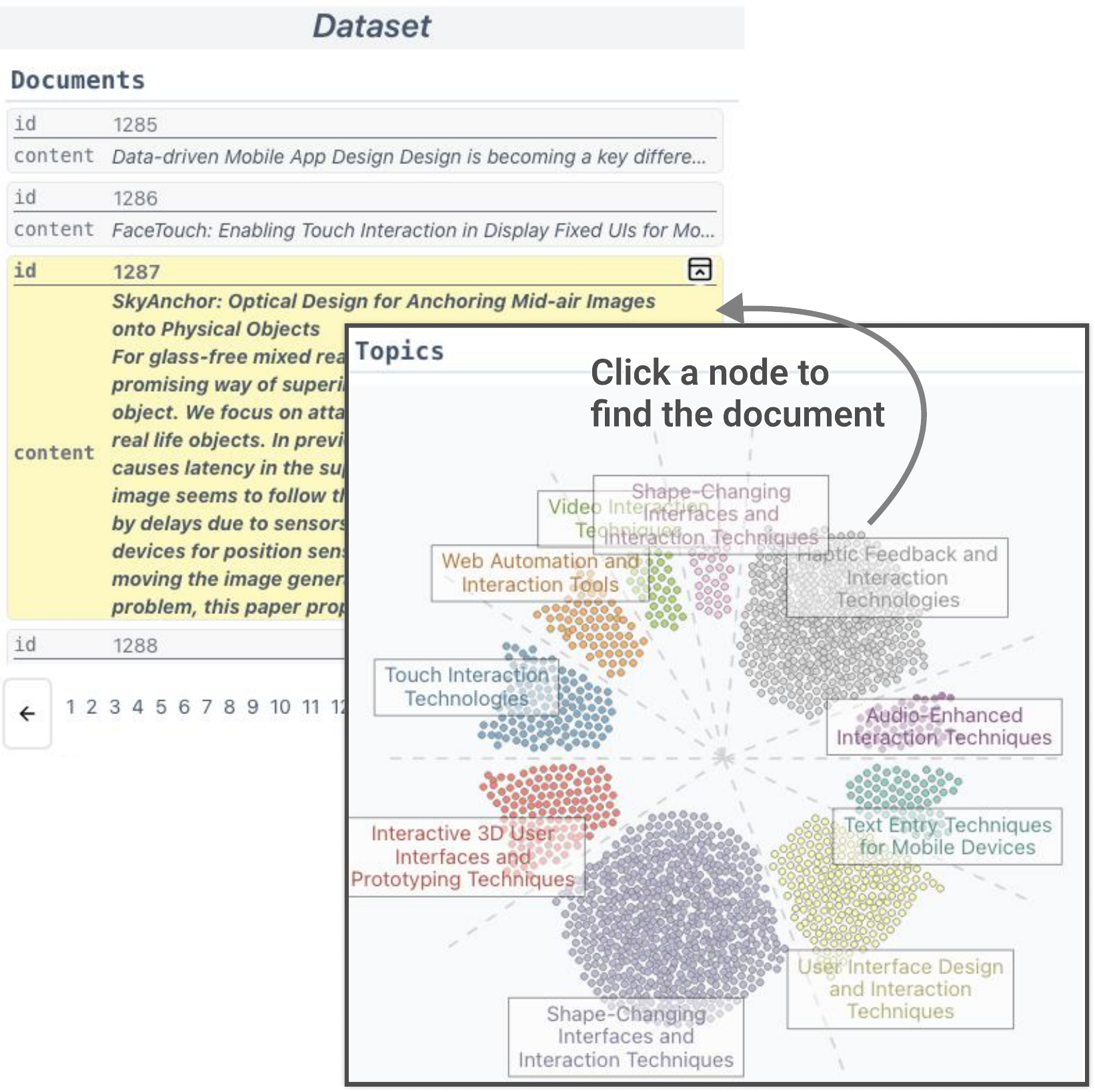}
    \caption{Demonstration of Data Inspection View using a HCI paper abstract dataset. Left: a simple list view showing the raw content of  documents. Right: the topic radial chart plotting the topic distribution of the documents. Each node is a document and each fan-shape region is a topic, such as ``Shape-Changing Interfaces and Interaction Techniques''. Clicking a node in the topic radial chart will highlight the corresponding document in the list view. 
    The topic radial chart faciliates the sensemaking process of text analytics.
    }
    \vspace*{-0.5cm}
    \label{fig: dataset-inspection}
\end{figure}

Sensemaking of the dataset is a critical step in text analytics, and is supported by the Dataset Inspection View through a paged document list and a topic radial chart (\autoref{fig: dataset-inspection}).

Builing on prior works on support sensemaking with topic models~\cite{grootendorst2022bertopic, lam2024lloom}, we design the topic radial chart as the basis for dataset inspection and result evaluation.
The topic radial chart divides the canvas into fan-shaped regions that represent topics.
Each document is represented as a circle and placed in the corresponding region. 
The topic radial chart can further encode the distribution of a numerical attribute associated with each document, using its distance to the center of the canvas. 
While not used in Dataset Inspection View, the same topic radial chart with distance encoding is used during the Evaluation stage (\autoref{fig: evaluation_results}), ensuring visual continuity for users from sensemaking to evaluation.

Being an instance of unit visualization~\cite{park2018unitvisualization}, the topic radial chart inherits many benefits of the unit visualization family.
The relative size of each topic can be visually estimated by the area or the angle of the topic region, and the absolute size can be estimated by the number of circles in the region.
The chart also supports interaction on individual documents: Clicking on a circle triggers the list view to automatically navigate to the clicked document.
Finally, the chart compactly encodes both the topic distribution and the evaluation score distribution.

\vspace*{-0.1cm}
\subsection{Plan View} 
In the second page, the Plan View shows the plan selected from Searching View. 
Here, the system also shows dependencies between semantic tasks that were hidden in previous page for visual clarity. 
Users can make final adjustments to the semantic tasks before execution, and click the ``Convert'' button to trigger the executor agent, which generates a pipeline of \textit{primitive tasks}.
Each semantic task is linked to the corresponding primitive task in Execution View.

\vspace*{-0.1cm}
\subsection{Execution View} 
Similar to semantic tasks, each primitive task has a label, description, and explanation. 
Each primitive task is selected from a pre-defined list consisting of well-established text analysis tasks, such as entity extraction, embedding generation, clustering, and dimensionality reduction. 
\subsubsection{Compilation}
As the user inspects the primitive tasks, the executor agent automatically starts the \textit{compilation} process, which generates technical details such as input/output schema, algorithm choices, and hyperparameters (details introduced in~\autoref{sec: backend_exexutor}), which typically takes less than 10 seconds.
Once the pipeline is compiled, users can see the input output schema and parameters by clicking the ``Inspect'' button.
Then, users can make any adjustments on the information shown in the inspection view (\autoref{fig: execution-evaluation-overview}-d).

\subsubsection{Execution and Results}
After compilation, a primitve task can be executed using the ``Execute'' button. The executor agent will automatically fetch the data, run the specified functions, and collect the results. 
The results can be inspected in their raw form in the Inspection View (\autoref{fig: execution-evaluation-overview}-e).
This inspection of results allows users to perform a quick sanity check to identify obvious errors. 
Users can edit and re-execute each node to refine the pipeline.

\subsection{Evaluation View}\label{sec: interface_evaluation}
Evaluation of a primitive task can be triggered once execution is complete. 
Currently, the system only supports LLM judges as evaluations, which only applies to tasks that generate textual responses.
Methods such as dimensionality reduction or cluster analysis requiring specific sets of evaluation metrics are not supported. 
Due to concerns that LLM judges may not accurately estimate numbers~\cite{chiang-lee-2023-large}, we restrict LLM judges to always output categorical scores.  
\subsubsection{Creating an Evaluator}
Users can click the ``plus'' button on the top right of Evaluation view, select a primitive task as the target, and describe the evaluation criterion in natural language.
The evaluator agent will automatically generate a prompt template for LLM judges using the description. Users can inspect the prompt template and make changes.
Additionally, the evaluator agent automatically recommends $k$ evaluation criteria for each execution node that generates textual responses. We empirically set $k=3$ to cover a reasonable amount of criteria while not \mbox{overwhelming the user}. 

\subsubsection{Result Visualization}
\begin{figure}[t]
    \centering
    \includegraphics[width=\columnwidth]{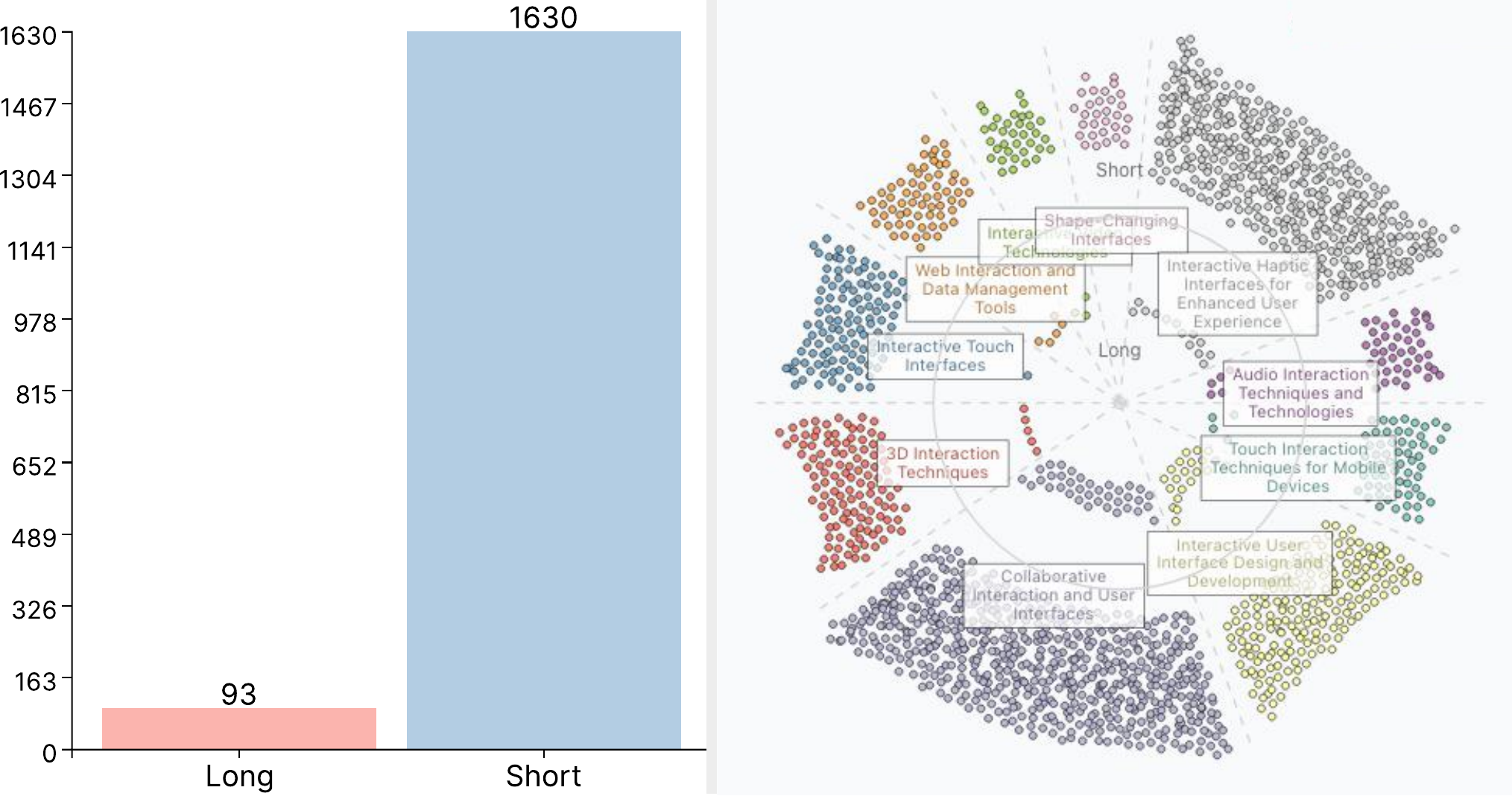}
    \caption{Visualizations for the evaluation results of ``Summary Length Evaluator''. Left: an overview of the distribution of categories in a bar chart. Each bar is a category generated by the LLM judges, and the height encodes how many units (e.g., documents) are assigned this category (Long or Short). Right: The topic radial chart extended to encode the categories. Each topic region is further subdivided by the categories. The inner region represents ``Long'' documents, and the outer region represents ``Short'' documents. 
    The visualization ensures visual continuity between sensemaking of the dataset and evaluation of execution results. 
    }
    \label{fig: evaluation_results}
    \vspace*{-0.5cm}
    
\end{figure}

The evaluation results are visualized in two ways (\autoref{fig: evaluation_results}).
Since the outputs of LLM judges are always categorical, bar charts are used to show the category distribution 
If the evaluation is run on document content, an extended version of the radial topic chart (introduced in \autoref{sec: dataset_inspection_view}) is used.
The extension divides each topic region into subregions, each representing a category. The documents, represented as circles, are placed in the corresponding subregion.
Since regions closer to the center have smaller areas and are more likely to be cluttered, the system assigns regions by category frequencies, where higher frequency categories are assigned with larger region, while maintaining the semantic order of the category (e.g., Low, Medium and High). 
This extension inherits most of the visual designs when users are making sense of the datasets, ensuring the visual continuity between evaluation and sensemaking.

\section{Backend Architecture}\label{sec: system_architecture}
We introduce three system modules---Decomposer, Executor, and Evaluator---that supports the corresponding agents, and a conceptual framework for analysis units to manage data structure changes.
We discuss the algorithmic choices, technical considerations, and implementations. 
\subsection{Specification Language} 
We defined a JSON-based specification language tailored to text analytics to bridge agent generations and user interactions.
The specification language describes the format for semantic tasks, primitive tasks, and evaluator nodes. 
The agents generate responses in the format of the specification language, while user interactions in the interface trigger modifications in the language, enabling the system to support human interventions. 

 \vspace*{-0.3cm}
\subsection{Decomposer}\label{sec: system_decomposer}
To efficiently and comprehensively search the decomposition space, the decomposer agent follows a generative reasoning approach~\cite{xie2023beamsearch} to iteratively search for alternative next steps of the plan. 
\begin{figure}[b]
    \centering
    \includegraphics[width=\columnwidth]{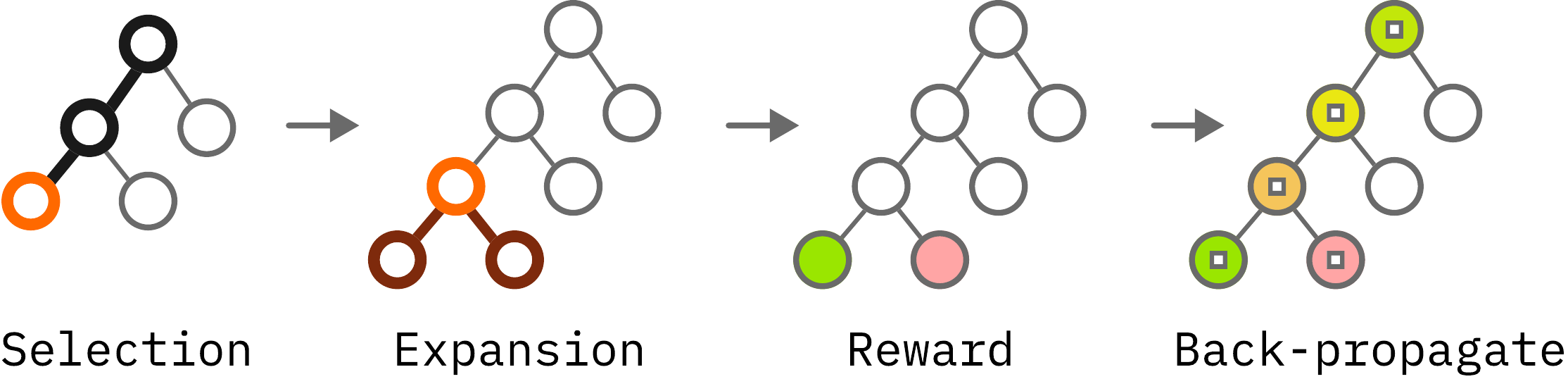}
    \caption{The process of Monte-Carlo Tree Search (MCTS) algorithm. Each node is a semantic task and each path represents an analytics plan. At each iteration, a node is selected with a seleection strategy (e.g., UCT selection or Greedy selection). Then, the selected node is expanded with $k=2$ children. For each child, we calculate a reward value using an ensemble of LLM judges, each ensemble has three scoring criteria: complexity, coherence, and importance. The reward is back-propagated until the root node is reached. 
    }
    \label{fig: MCTS}
    \vspace*{-0.5cm}
\end{figure}
\subsubsection{Search Algorithm}
The decomposer module implements a Monte-Carlo Tree Search (MCTS) algorithm with several modifications to adapt to generative reasoning and support human intervention.
Previous work has combined beam search with self-evaluation rewards~\cite{xie2023beamsearch}, in which each reasoning step is treated as a token in a sequence and rewards are assigned by LLM judges.
However, the greedy maximization of beam search is not ideal for text analytics pipelines since the value of a step can not be accurately estimated until later steps are introduced.  
In contrast, the backpropagation in MCTS provides a backward feedback mechanism (\autoref{fig: MCTS}).
At every iteration, the algorithm expands a node with $k=2$ children as alternative next steps. 
For each new child, the reward is calculated by a committee of LLM judges and backpropagated to the root node.
This way, the reward of previous nodes is adjusted by the reward of later nodes, effectively providing a backward feedback.
The full details of the MCTS algorithm are introduced in the Supplemental Materials. 

\subsubsection{Reward Function}\label{sec: reward}
We use a committee of LLM-judges as the reward function in the search process. 
Compared to self-consistency-based evaluation methods~\cite{wang2023selfconsistency}, which aggregate model outputs based on agreement levels, LLM judges provide a more context-aware and adaptable mechanism~\cite{beigi2024uncertaintyreview}.
Discussions among the co-authors converged to three criteria: complexity, coherence, and importance, each rated by an LLM judge, as defined in~\autoref{tab:decomposition_criteria}.
\begin{table}[htbp]
\centering
\begin{tabular}{p{2cm} p{6cm}}
\toprule
\textbf{Criterion} & \textbf{Description} \\
\midrule

\textbf{Complexity} &
Assesses whether a task can be executed as a single step in a pipeline. 
During decomposition, each step should not be further divisible and must 
be convertible into an executable operation. This criterion promotes 
simplicity and facilitates efficient execution. \\
\hline
\textbf{Coherence} &
Evaluates the logical consistency between consecutive steps. A coherent 
sequence ensures that each task logically follows from its predecessor, 
avoiding contradictions, redundancies, or irrelevant operations. This 
maintains a meaningful analytical flow. \\

\hline
\textbf{Importance} &
Measures the contribution of each task to the overall analytical goal. 
Each task should be critical and essential to achieving the final outcome. 
Optional or peripheral tasks are excluded to maintain focus and execution efficiency. \\

\bottomrule
\end{tabular}
\vspace{0.1cm}
\caption{Criteria for evaluating task decomposition quality.}
\label{tab:decomposition_criteria}
\vspace*{-0.5cm}
\end{table}




To ensure reward function reliability, we incorporate several technical considerations.
First, the LLM judges produce likert scale scores rather than open-text~\cite{chiang-lee-2023-large}.
Second, each criterion is rated by three different models, taking average as the final score.
Third, the reward of a reasoning path is computed using the geometric mean of the node rewards, balancing both local and global quality while penalizing weak links. 

\subsubsection{Taking user feedback}
Users can manually revise scores and optionally provide explanations for their decisions. These explanations are stored as few-shot prompts to adjust the LLM judges’ behavior in subsequent evaluations. Users can also redefine the scoring criteria, enabling dynamic and domain-specific evaluation strategies.

\subsubsection{Implementation}
The decomposer is implemented using the AutoGen~\cite{wu2023autogen} framework for coordination among multiple LLM judges. 
The three models for reward scoring are GPT-4o~\cite{openai2024gpt4}, Claude-3.5-Sonnet~\cite{anthropic_claude}, and Gemini-2.0~\cite{gemini_model}. 


 \vspace*{-0.3cm}
\subsection{Executor}\label{sec: backend_exexutor}
Given the semantic tasks, the executor creates an executable pipeline in two steps: \textit{Conversion} and \textit{Compilation}.

\subsubsection{Conversion}
The executor agent converts semantic tasks into primitive tasks during conversion.
For each semantic task, the agent selects the corresponding primitive tasks from a predefined list based on text analytics literature~\cite{Liu2019textminingsurvey}.
Each item in the list contains the definition and input/output data (e.g., text or embeddings), which helps ensure data structure compatibility between consecutive primitive tasks.
The full list of primitive tasks is in the Supplemental Materials.

\subsubsection{Compilation}
Based on the conversion result, a complete execution specification is generated in the following procedure:
First, the \textbf{input/output schema} is generated to specify how to manage the data structure of the results.
The generation process follows a conceptual framework of \textit{Analysis Unit} to infer the correct input/output schema, which is introduced in~\autoref{sec: handling schema}.
Next, the necessary \textbf{tools and parameters} for each primitive task are generated.
The supported tools (i.e., function calls) cover most common text analytics needs, such as prompt template tool, clustering tool, and data transformation tool.
For tools requiring code execution (e.g., clustering), specialized prompts are developed to pick a specific algorithm (e.g., K-Means), and hyperparameters.
Finally, the \textbf{execution graph} is constructed and stored using the specification language to allow human intervention.
Before executing a primitive task, the executor gathers all the compiled information to create a LangGraph~\cite{langgraph} executable.
\subsubsection{Human Intervention}
\system\ supports (1) \textit{Modifying task definition} (labels and descriptions), (2) \textit{Adjusting input/output schema} to align intention, e.g., changing from ``topic'' to ``theme'', and (3) \textit{Tuning execution parameters}, such as editing prompt templates and algorithm hyperparameters.

\rev{
\subsubsection{Parallelization and implementation} 
\marginnote{$\triangle$\_6\_1}
The system leverages parallelization of cloud LLM requests to optimize executions and integrates with LangGraph~\cite{langgraph}. 
}

\vspace*{-0.4cm}
\subsection{Conceptual Framework for Analysis Units}\label{sec: handling schema}
\begin{figure}[t]
    \centering
    \includegraphics[width=\columnwidth]{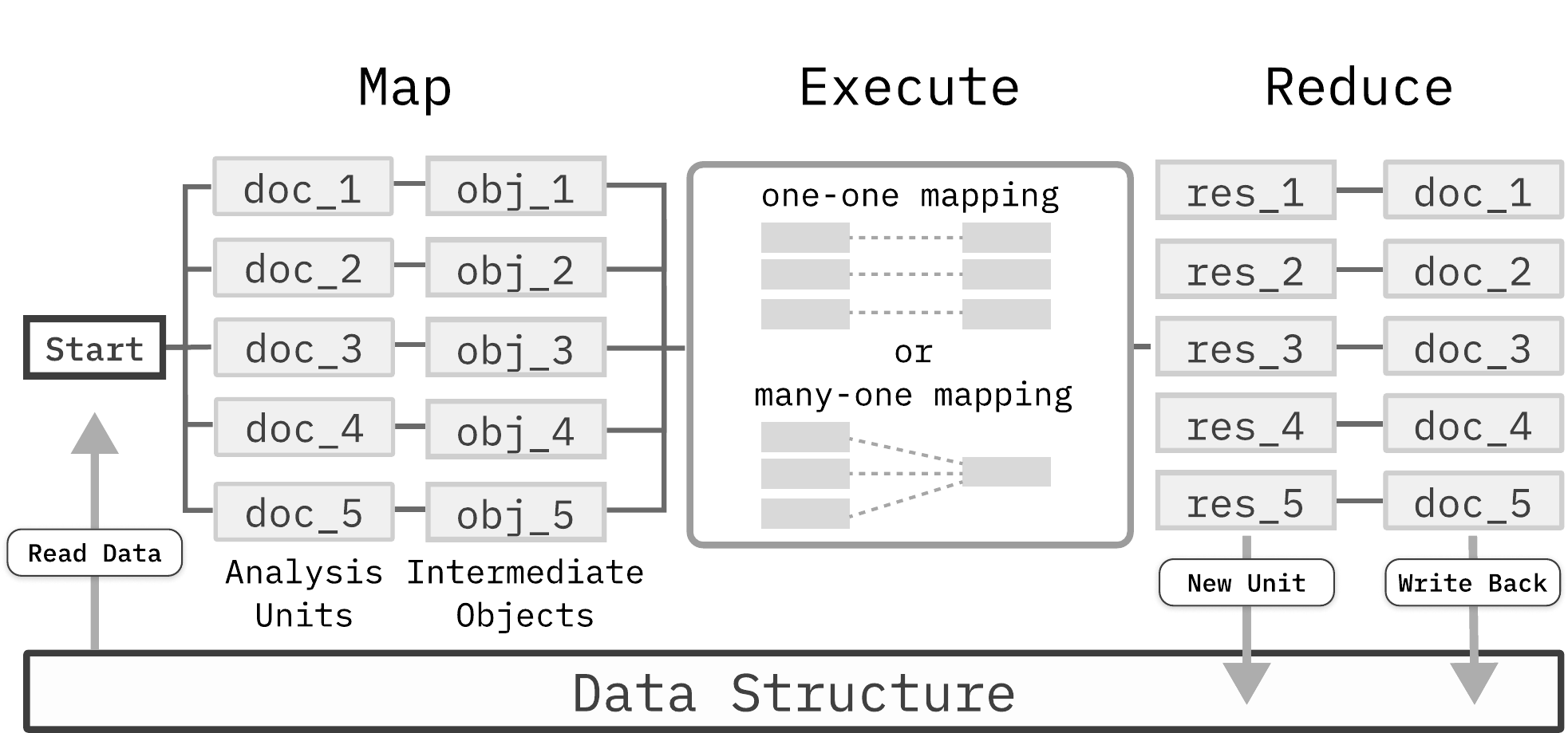}
    \caption{The conceptual framework that defines how each primitive task handles data structure changes. First, data is read from a data structure that stores all the results. Each analysis unit (e.g., document) is mapped to an intermediate object that are the direct inputs of an execution function. The Reduce step writes the execution results back to each input unit, and additionally creates a new analysis unit adds it back to the data structure. 
    This framework ensures all later steps can access the new units created by previous steps.
    }
    \label{fig: analysis_unit}
    \vspace*{-0.5cm}
\end{figure}

Starting with unstructured documents, a text analytics pipeline enriches the data structure after each primitive task.
To automatically generate a pipeline, it is critical to explicitly manage the data structure to allow later tasks access to new structures. 
To this end, the compilation agent follows a conceptual framework of \textit{Analysis Unit}, as shown in~\autoref{fig: analysis_unit}. 

\subsubsection{Analysis Units} 
The analysis unit defines \textit{what} serves as input to a primitive task. As the pipeline progresses, new units may be created. For example, in entity extraction from documents, each \textit{document} starts as a unit, and the extraction step generates a new unit, \textit{entity}, for subsequent steps.


\subsubsection{Generating executable with MapReduce}
The executable code of each primitive task is generated with the MapReduce paradigm: (1) \textbf{Map}: Given a list of analysis units, create a list of intermediate objects that one-to-one map to the units. For example, from a list of documents, create a list of objects that contains the content of each document. (2) \textbf{Execute}: On the mapped result, execute a function. The function can target individual objects or a group of objects, e.g., a function that extracts entities from each document or groups the documents by topics. (3) \textbf{Reduce}: Store the execution result back to the data structure with necessary aggregation operations. 


\subsubsection{Automatic Unit Creation} 
Combining the MapReduce paradigm and analysis units, the system automatically creates new units after executing a primitive task. 
In the reduce step, in addition to storing the execution results, the system assumes a new unit is created and aggregates the results to represent the new unit.
For example, after entity extraction on documents, we not only store the extracted entities of each document, but also create a new unit named \textit{``entities''}, storing all entity occurrences. The new ``entity'' unit is no longer under ``document'', and can be accessed by subsequent tasks.

\subsection{Evaluator}
The evaluator module implements the functionalities needed to support the evaluation of execution results.

\subsubsection{The No-Ground-Truth Assumption}
One critical consideration when designing the evaluation is the ``No-Ground-Truth'' assumption, i.e., no human annotations or labels are available.
Prompt engineering with optional few-shot examples has replaced the conventional way of collecting training samples to improve model performance, as collecting ground-truth labels for prompt templates is impractical.

As a result, conventional metrics such as F1 score are not applicable. 
While, researchers are still exploring new evaluation methods~\cite{beigi2024uncertaintyreview}, LLM judges are currently the most widely adopted evaluation method~\cite{Shankar2024evalgen}. 
LLM judges are typically used to evaluate \textit{local} characteristics, i.e., evaluations are conducted per analysis unit.
For example, the textual features of summarizations~\cite{Lee2025Awesum}, which are calculated per generated summary, or the language simplicity and scientific accuracy of scientific explanations~\cite{kim2024evallm}. 
Although limited, these local characteristics evaluated by LLM judges facilitate users to quickly make sense of a large collection of responses. 




\vspace*{-0.3cm}

\section{Quantitative Experiments}
Technical evaluations are conducted on the decomposer and executor module.
For decomposer, we compared the generated pipelines against human-crafted pipelines.
For executor, we use a human-crafted pipeline with generated execution parameters and evaluate the results on a dataset with ground truths.

\vspace*{-0.3cm}
\subsection{Decomposer Evaluation}
We evaluate the performance of the decomposer agent on two usage scenarios: LLooM~\cite{lam2024lloom} and TnT-LLM~\cite{Wan2024TnTLLM}.
Each scenario involves a text analytics pipeline with a realistic goal (\autoref{tab:pipeline_goals}). Our experiment compares the pipelines generated by the decomposer agent without human intervention with those defined by the authors of LLooM and TnT-LLM. 


\begin{table}[htbp]
\centering
\begin{tabular}{p{2cm} p{6cm}}
\toprule
\textbf{Pipeline} & \textbf{Goal} \\
\midrule

\textbf{LLooM} &
\textit{Extract high-level concepts to understand research topics 
from a dataset of HCI paper abstracts.} \\

\addlinespace
\textbf{TnT-LLM} &
\textit{Understand the user's intents when interacting with the chatbot 
from a dataset of user conversations with the Microsoft Bing Consumer Copilot system.} \\

\bottomrule
\end{tabular}
\vspace{0.1cm}
\caption{Analytical goals of the evaluated pipelines.}
\label{tab:pipeline_goals}
\vspace*{-0.5cm}
\end{table}


\subsubsection{Creating Pipelines}
For each paper, two pipelines are created and compared.
The first pipeline, \textit{manual pipeline}, recreates the pipeline described in the corresponding paper using our specification language.
For example, the TnT-LLM pipeline is specified in four steps: \textit{Summarization}, \textit{Data Transformation} (pick a subset of summaries), \textit{Label Generation} (produce a label taxonomy from the subset of summaries), and \textit{Document Classification}.
The second pipeline, \textit{generated pipeline}, consists of primitive tasks generated from the decomposer agent.
The decomposer agent first run the MCTS algorithm until a complete tree is reached (all leaf nodes are ``END'' nodes), and then pick the path with the highest value.

\subsubsection{Evaluation Method}
Our requirement for generated pipelines is to be \textit{on par with} manual pipelines, as we do not require the pipelines to be superior to the expert-crafted ones.  
Therefore, our evaluation method follows the Arena~\cite{chiang2024chatbotarena} approach, i.e., two pipelines are anonymously presented for evaluators to pick a better one.
We leverage state-of-the-art reasoning model (OpenAI o3-mini) as evaluators with clear criteria.
If the reasoning model does not consistently pick the manual pipeline as the better one, we conclude that the generated pipelines are \textit{on par with} the manual pipelines. 
This process is repeated five times for each pipeline comparison with randomized orders to avoid position bias~\cite{shi2024llmjudgespositionbias}, resulting in five pipelines per paper and ten comparisons.
Prompts and intermediate results are in the Supplemental Materials.

\subsubsection{Results and Analysis}
Out of 10 comparisons, 6 of the generated pipeline are considered better (2 for LLooM and 4 for TnT-LLM).
In general, the model recognizes our generated pipelines are of comparable quality to human-crafted ones.
To provide a deeper understanding of the selection, we summarize the overarching pros and cons of the generated pipelines below.

\paragraph{Direct and Streamlined Solutions}
The generated pipelines are praised for offering a direct and streamlined solution to the goal, while manual pipelines generally have extra technical decisions. For example, the manual pipeline for LLooM first summarizes each document into a list of bullet points, then conducts cluster analysis on the bullet points, yet the benefit of summarization is not apparent. 
The generated pipelines might be more direct and streamlined due to the influence of coherence and importance scores in path selection.

\paragraph{Failure to Consider Long Data}
The manual pipelines are often favored for the consideration of handling long input. To prevent LLM context window overflow, both manual pipelines have extra summarization steps to reduce input length. The generated pipelines often fail to consider such limitations, which we speculate is due to the lack of explicit knowledge of context window limitations. 

\rev{
\subsubsection{Cost Analysis}
\marginnote{$\triangle$\_7\_1}
On average, the MCTS algorithm takes 5 seconds per expansion, including one LLM request to generate two children and 18 ($2\ children \times 3\ criteria \times 3\ judges$) parallel LLM requests. 
The average cost of API usage per expansion is 0.005\$. 
On average, a complete tree has 80 nodes and takes up to 7 minutes to generate.
}

\subsection{Executor Evaluation}
We evaluate the executor by recreating the LLooM pipeline~\cite{lam2024lloom} and comparing the performance with LLooM.
We recreate the pipeline of LLooM using our system with five primitive tasks: \textit{Summarization}, \textit{Embedding Generation}, \textit{Clustering}, \textit{Data Transformation}, and \textit{Label Generation}.
The prompt templates and parameters for each step are automatically generated using the compilation procedure. 

\subsubsection{Dataset and Metric}
We use the Wikipedia dataset~\cite{pham2024topicgpt} for its ground-truth labels for topics. For evaluation metric, we use \textit{concept coverage} as defined in LLooM, which measures the proportion of ground truth topics that are covered by the generated concepts. The metric uses a prompt template to identify covered concepts. In our experiment, we use the same prompt as LLooM, but run the prompt on GPT-4o to keep up with the advance in foundation models.


\subsubsection{Procedure}
BERTopic~\cite{grootendorst2022bertopic} and GPT-4o are baselines to compare performance. 
Following the method in LLooM, we first use random samples of the dataset documents (n=210) stratifed across topics to accommodate context window limits.
Then, we execute the baseline methods and the pipeline on the sampled dataset for ten iterations.

\begin{figure}[h]
    \centering
    \includegraphics[width=\columnwidth]{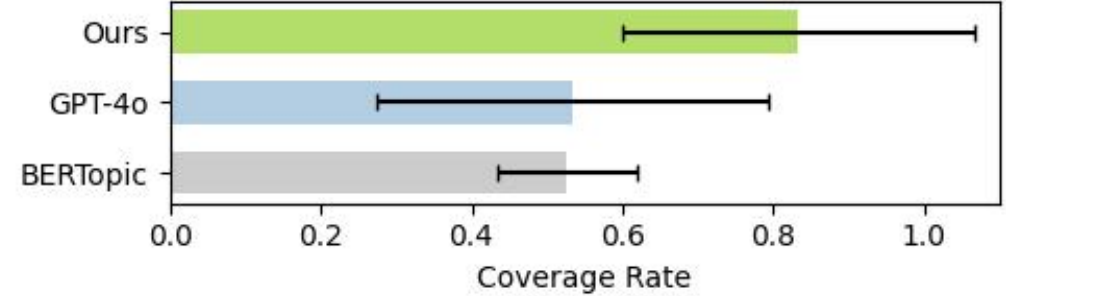}
    \vspace*{-0.5cm}
    \caption{Results of the executor experiment. On the Wikipedia dataset, our pipeline with generated parameters exceeds baseline performance achieving a 83\% coverage rate (52\% for BERTopic, 53\% for GPT-4o). }
    \vspace*{-0.2cm}
    \label{fig: execution_experiment_result}
\end{figure}
\subsubsection{Results}
As shown in~\autoref{fig: execution_experiment_result}, the pipeline exceeds baseline coverage rate by 30\% (BERTopic = 52.6\%, GPT-4o = 53\%, ours = 83\%), suggesting that the pipeline achieves comparable results with the human-crafted pipeline in LLooM.
This shows the reliability of using LLM-generated prompts and execution parameters to replace human-crafted ones.

During the experiment, we find that the executor sometimes picks the wrong analysis unit for a primitive task during compilation.
For example, consider a pipeline with two steps: Clustering Analysis and Label Generation (for the clusters). 
For the second task, Label Generation, the correct analysis unit is the \textit{cluster}, but the executor might pick \textit{document} as the analysis unit. 
These errors are difficult to detect and recover, and currently can only rely on human intervention to address.


\section{User Study}
To evaluate the usability and utility of \system, we conducted a qualitative user study on participants with varying levels of experience in text analytics.

\vspace*{-0.3cm}
\subsection{Study Design}
\subsubsection{Recruitment and Participants}
We recruited six graduate students in the field of NLP or data science as participants.
All participants have experience in data science with a passing knowledge of data analysis.
The participants are divided into three groups according to their experience in text analytics: 
\begin{itemize}[leftmargin=4mm]
    \item \textbf{No experience} with NLP models or text analytics (\textbf{{P1, P2}}).
    \item \textbf{Some experience} using NLP models or building text analytics pipelines (\textbf{{P3, P4}}).
    \item \textbf{Expert-level experience}, who participated in research projects using NLP models or text analytics (\textbf{{P5, P6}}).
\end{itemize}
Our study was determined to be exempt from review by our Institutional Review Board (IRB), and we obtained informed consent from all participants prior to the study.
\subsubsection{Task Design}
Participants are given the \textit{Concept Induction} task as defined in LLooM~\cite{lam2024lloom}: \textit{Generate high-level concepts from a corpus}, on a dataset with 100 HCI paper abstracts~\cite{cao2023HCIdataset}.
No participants have prior knowledge of HCI. 
We divide the task into two subtasks with an optional break considering the cognitive demand to conduct end-to-end text analytics.
The first subtask requires participants to complete \textit{Decomposition} using the Searching View. 
The second subtask requires participant to complete \textit{Execution} and \textit{Evaluation} using the Execution and Evaluation View. 
To test whether the interactions support users in identifying errors, we picked common errors occurred during quantitative experiments and placed them into a pipeline.
Participants are instructed to find and fix the errors while executing and evaluating the pipeline. 

\subsubsection{Procedure}
The user study began with an introduction session that explains the research background and the interface, which typically took 30 minutes.
Then, we introduce concept induction and the HCI paper dataset. Participants were instructed to use the interface freely to complete the two subtasks with the think-out-loud protocol in 30-45 minutes. 
After the tasks, we conducted a semi-structured interview to gain a deeper understanding of the user experience.
On average, the user study took 1 hour and 45 minutes and participants were compensated \$20 for their time.
We recorded the screen and audio and transcribed them for further analysis.

\vspace*{-0.3cm}
\subsection{Usability and Utility Results}

\subsubsection{Clear and straightforward workflow}
\rev{
\marginnote{$\triangle$\_8\_1}
Overall, the user study yielded positive results, with all participants completing the tasks within expected time. The clear and structured workflow was praised for simplifying a complex analytical goal, especially for entry-level participants. The visualization design, particularly the searching tree, externalized the task space and progressively revealed semantic and primitive tasks, helping participants form accurate expectations and plan with reduced cognitive effort. As P1 noted, ``\textit{(the searching tree) allows me to explore more tasks that I would not think of myself... and I feel prepared before moving on to execution.}'' These results highlight the value of visualization in scaffolding analytics for entry-level data analysts.
}
\subsubsection{Significance of automation}
Participants all agree that conducting text analytics using \system \ is more efficient than coding, even for expert-level participants.
The system is helpful in automatically handling some tedious and error-prone implementation actions, such as integrating execution functions in a script, writing prompt templates, or ensuring consistent input/output schema. Although all of this can be done manually, it is more efficient and robust to systematically connect code snippets into a coherent pipeline.

\subsubsection{Learning Process}
\rev{
\marginnote{$\triangle$\_8\_2}
Participants initially reported high cognitive demand due to the tightly integrated features. However, the visualization clarifies the workflow logic and supports mental model formation, easing the learning process. As a result, the system was learnable within a 30-minute introduction and training session, even for novices: ``\textit{Once you get the logic, it becomes easier and natural to use (P1)}.''
}
\vspace*{-0.3cm}
\subsection{Findings}
The user study uncovers several interesting behavior patterns among the participants, offering valuable insights into the dynamics of human-agent collaboration in text analytics.

\subsubsection{Searching strategy}
The typical MCTS algorithm balances between exploitation and exploration.
While the balanced strategy may be suitable for automatic searching, it is not necessarily desirable in the human-in-the-loop setting.
Most participants prefer the ``exploit first, explore later'' strategy, i.e., first complete searching on one best path, then optionally explore the alternatives based on the best path.
This strategy avoids overwhelming the users with too many options.

\subsubsection{Alignment vs. Suggestion}
Participants felt ambivalent about LLM judges and the feedback mechanisms in the Decomposition stage.
While they appreciate the feedback mechanism that enables alignment, they do not find it necessary to provide feedback.
This is because LLM judges are following explicit criteria that are perceived to be consistent, which improves their confidence in reward values. 
Participants prefer taking the scores as \textit{suggestions} rather than \textit{ground truths} and actively \textit{align} them~\cite{Shankar2024evalgen, liu2023calibratingllmbasedevaluator}.
This observation reveals that LLM judges do not necessarily need to be perfectly aligned with human as long as sufficient human autonomy is provided.

\subsubsection{Analysis units help understanding the pipeline}
Understanding a pipeline is a cognitively demanding process, and making the analysis units explicit helped.
By making sense of the changes of analysis units, participants can easily keep track of the execution progress, anticipate results, and identify errors. Many participants started debugging by mentally comparing the expected analysis unit of the input and output for each primitive task, and eventually found the cause of errors without external support.


\subsubsection{Gaining trust through procedural generation}
We find that participants reported more trust in the results generated by \system \ than other GenAI systems such as ChatGPT.
In comparison, most GenAI systems generate answers from a black-box decoding process, while participants prefer results generated from an explicit procedure, i.e., the text analytics pipeline.
As commented by P2: ``\textit{because it has some step-by-step process, it looks more convincing... and I am more confident in the results.}'' 


\subsubsection{Comparison of execution results}
Both expert-level participants (P5 and P6) expressed the desire to compare searched paths based on execution results. 
This suggests an extension of the linear three-stage workflow with backward feedback, in which the results from Execution and Evaluation inform the search process in the Decomposition stage.
By incorporating this feedback loop, users could refine the decomposition using the outcomes as evidence,
as opposed to using the purely deductive scores generated by LLM judges.





\section{Discussion}
\system\ is specifically designed for text analytics, but it addresses technical challenges common to agentic systems. In this section, we discuss the implications for text analytics, agentic systems, limitations of the system, and future work.

\vspace*{-0.3cm}
\subsection{Implications for Text Analytics}
\subsubsection{Democratizing Text Analytics}
Text data is arguably the most accessible data for the general public, yet the ability to analyze text data remains technically advanced even for data analysts~\cite{Liu2019textminingsurvey}. 
While individuals may feel confident expressing opinions, crafting arguments, or summarizing information in writing, they often lack the tools or knowledge required to systematically explore patterns, trends, or insights in textual data at scale. 
\rev{LLM-based agentic systems
significantly reduces the technical barriers~\cite{Subramonyam2024gulfofenvisioning}, indicating strong potential for democratizing text analytics~\cite{lefebvre2021datademocratization}.
Experiment results show current coding agents still require human oversight, and} \system \ is a first step towards this future, offering a user-friendly interface that simplifies the complexities of text analytics while ensuring the reliability and validity of results.
While \system \ exhibits a steep learning curve that is not suitable for the general public, it can benefit a wide range of professionals who have strong desire to analyze text data, such as journalists and social science researchers~\cite{drápal2023llmforlegal}.
We envision a future where more people can leverage text data to explore, understand, and analyze information.

\subsubsection{Balancing Rigor and Usability \rev{through Visual Design}}
While democratizing text analytics seems to be a desirable future, it is also critical to avoid issues such as overreliance on AI~\cite{passi2022overreliance}. 
Relying too heavily on autonomous agents without understanding the underlying processes can lead to oversimplification and erroneous conclusions. Therefore, users must be aware of alternative methods and avoid false findings.
To support such a process, the system must balance between technical sophistication and human interpretability, creating a dilemma for the system design. 
On the one hand, the system needs to enforce certain engineering practices to ensure the reliability of the results.
On the other hand, the system must also remain user-friendly, support intuitive interaction, minimize cognitive overhead, and enable critical reasoning and insight generation.
\rev{
\marginnote{$\triangle$\_9\_1}
\system \ shows that appropriate visual design is key to maintaining this balance. Representing plans and pipelines as trees and node-link diagrams and mapping manipulation of these visual elements to back-end operations prevents users from being overwhelmed by the complex technical operations to oversee and control agents. 
}

\vspace*{-0.3cm}
\subsection{Implications for Agentic Systems}
\rev{
\subsubsection{Visualization in Agentic Systems}
\marginnote{$\triangle$\_9\_2}
Current agents are implemented using non-deterministic foundation models under the ReAct~\cite{yao2023react} paradigm; however, our preliminary experiments suggest that this approach may be unreliable in complex scenarios.
\system\ decomposes text analytics into multiple stages, introducing a structured procedure with a clear separation of responsibilities between the human and the agent.
This procedural structure naturally enables shared representations~\cite{heer2019sharedrepresentation}, allowing the human and the agent to operate on the same objects.
In \system, the shared representation consists of a search tree and a node-link pipeline visualization, ensuring that all agent \textit{actions} are visible, predictable, and controllable to the user.
Although this collaborative paradigm introduces additional cognitive and metacognitive demands~\cite{tankelevitch2024metacognitive}, human oversight and control substantially improve system reliability~\cite{passi2025agentic_ai_human_oversight}, outweighing the added burden.



}
\subsubsection{LLM Alignment and Trust}
Practitioners still have concerns about LLM judges~\cite{Shankar2024evalgen} even when alternative approaches perform worse. 
In \system, LLM judges are heavily used, but participants did not feel the need to perform alignment.
We argue that this is due to the role dynamic between human and agent: the users see themselves as managers and perceive the LLM judges as consultants, i.e., the comments of consultants are helpful, but the manager makes the final decision.
Contrary to other usage contexts, where LLM judges are treated as replacements for ground truths,
our human-agent dynamic facilitates the metacognition~\cite{tankelevitch2024metacognitive} process in GenAI systems.
Our findings reveal that simply changing the human-agent collaboration dynamic can reduce the necessity of LLM alignment and foster trust in agentic systems.

\vspace*{-0.3cm}
\subsection{Limitations and future work}
\rev{
\subsubsection{User study limitations}
\marginnote{$\triangle$\_9\_3}
Six participants is not a large sample size and the recruitment could be considered biased as all participants are graduate students. Also, usability is not quantitatively evaluated.
}
Despite these limitations, the participants are representative of our target users: entry-level data analysts in text analytics, and we discovered interesting findings regarding the human-agent collaboration workflow.
\subsubsection{Limitations of the system}
First, the system relies heavily on parallelized cloud-based LLM queries to maintain acceptable response time and interactivity, which is not transferable to local LLM queries. 
\rev{
\marginnote{$\triangle$\_9\_4}
Second, although the LLM judges in the system were acceptable to our participants, they have not been rigorously evaluated for alignment with human judgment and therefore should not be reused.
}
Third, the system lacks a robust self-correction mechanism for erroneous agent responses. 
Finally, the system does not incorporate any metadata, such as dates or sources of documents, nor does it connect to an external knowledge base, and thus, the system does not support time-series or network analysis and does not apply to domain-knowledge-intensive contexts.
We consider them to be out of the scope of our research objective.

\subsubsection{Future work}
Conversational agents that can translate user commands to changes on the text analytics pipeline can be incorporated to enhance usability. 
\rev{
\marginnote{$\triangle$\_9\_5}
The system could also be improved by connecting the results of execution and evaluation to the decomposition stage, and incorporating analysis units in the tree design.
}
For the evaluation stage, evaluation methods for tasks like cluster analysis or dimensionality reduction should be supported and integrated with recent work in agentic visualization~\cite{chen2025interchat, Chen2025Viseval, Tian2025ChartGPT}.
Moreover, \system \ can significantly improve the reproducibility of existing bespoke visual text analytics systems~\cite{Jänicke2017visualtextanalytics, Liu2019textminingsurvey} by recreating the text analytics pipeline proposed in the respective papers.
Finally, \system \ is highly extensible, as it is implemented with common open-source frameworks.

\section{Conclusion}
We build upon recent advances in agentic systems and develop \system, which allows human oversight and control of agents in text analytics. 
The system can be extended to incorporate conversational agents, more evaluation and visualization methods, and integrate with the open-source community.
The system sheds light on promising directions for democratizing text analytics, and we discover interesting insights regarding the impact of human-agent dynamics on the reliability of agentic systems and LLM alignment.

\vspace*{-0.3cm}
\subsection{Acknowledgments}
\noindent This research is supported in part by the National Science Foundation via grant No. IIS-2427770, and by the University of California (UC) Multicampus Research Programs and Initiatives (MRPI) grant and UC Climate Action Initiative grant.

\balance
\bibliographystyle{plain}
\vspace*{-0.1cm}
\bibliography{references}


\newpage
\onecolumn
\begin{multicols}{2}
\section{Monte-Carlo Tree Search Algorithm for Generative Reasoning}
In this section, we present the full technical modications made to the Monte-Carolo Tree Search algorithm (MCTS).
\subsubsection{Modifications of MCTS}
To adapt MCTS to generative reasoning, we make several modifications to the original MCTS algorithm. 
\paragraph{LLM judges as rewards}
To improve the reliability of LLM judges, we explicitly define three categorical scoring criteria (complexity, coherence, and importance) and use an ensemble of LLM judges.
The scores are aggregated to produce a continuous reward value. More details are introduced in~\autoref{sec: reward} 

\paragraph{Replacing random playouts}
Second, we calculate the reward for all children after each expansion. 
The original MCTS algorithm runs playout on a randomly selected child to improve efficiency. This is not necessary when the reward function is implemented by LLM judges, since we can efficiently calculate the rewards for all children. This also simplifies the backpropagation step for the user to understand, because all the rewards of the children are backpropagated after each iteration, instead of a randomly selected child. 

\paragraph{Human Intervention}
By integrating the MCTS results in the specification language, the search process can be paused for human intervention after each iteration. When resumed, the search continues on the current tree as defined in the specification language, which allows human intervention to be incorporated in subsequent iterations. 
\end{multicols}

\section{List of Primitive Tasks}
The following table shows the full list of primitive tasks that the system supports. This list is used in the conversion process of the  executor module.
\begin{longtable}{@{} 
    >{\raggedright\arraybackslash}p{0.15\textwidth} 
    >{\raggedright\arraybackslash}p{0.45\textwidth} 
    >{\raggedright\arraybackslash}p{0.1\textwidth} 
    >{\raggedright\arraybackslash}p{0.2\textwidth} 
  @{}}
    \toprule
    \textbf{Label} & \textbf{Definition} & \textbf{Input} & \textbf{Output} \\
    \midrule
    Entity Extraction & Finds and categorizes entities like names, places, and dates. & Text & Entity List \\
    \addlinespace
    Relationship Extraction & Finds relationships between entities in text. & Text & Relationship List \\
    \addlinespace
    Sentiment Analysis & Identifies the sentiment of the text as positive, negative, or neutral. & Text & Text (Sentiment Label) \\
    \addlinespace
    Document Classification & Assigns categories to a document based on its content. & Text & Category Label List \\
    \addlinespace
    Label Generation & Generates keywords or labels summarizing a document. & Text & Keyword Label List \\
    \addlinespace
    Summarization & Creates a shorter version of the text while keeping key information. & Text & Text (Summary) \\
    \addlinespace
    Embedding Generation & Converts text into numerical vector representations for analysis. & Text & Vector Representation \\
    \addlinespace
    Clustering Analysis & Groups similar items together based on their vector representation. & Vector Representation & Cluster Label List \\
    \addlinespace
    Dimensionality Reduction & Reduces the number of dimensions in vector representations while preserving important patterns. & Vector Representation & Vector Representation \\
    \addlinespace
    Data Transformation & Converts data into a different format or structure for analysis or modeling. & Any & Transformed Data \\
    \addlinespace
    Machine Translation & Translates text from one language to another. & Text & Translated Text \\
    \addlinespace
    Question Answering & Finds answers to questions based on a given text. & Text & Answer (Text) \\
    \addlinespace
    Natural Language Inference & Determines the relationship (e.g., entailment, contradiction) between two pieces of text. & Text & Inference Label (Text) \\
    \addlinespace
    Segmentation & Splits text into meaningful segments like sentences, paragraphs, or topics. & Text & Text Segment List \\
    \addlinespace
    Insights Summarization & Synthesizes key insights and findings from previous analysis results or multiple data sources. & Any & Text (Insights Summary) \\
    \bottomrule
\end{longtable}

\onecolumn
\section{Prompts}
\begin{multicols}{2}
In this section, we enumerate all the prompts that are used in the system.

(1). Generates semantic steps for a text analytics plan based on the goal and the prior progress. Used in the searching stage.

\begin{promptbox}
\noindent
** Context **

You are a text analytics task planner. 
Users have collected a dataset of documents. The user will describe a goal to achieve through some text analytics, and what they have done already.

\noindent
** Task **

Your task is to provide a single next step, but with {n} possible alternatives for user to choose.

Focus on the conceptual next step in terms of text analytics. If no further steps are needed, label the next step with "END". The plan should not be more than 5 steps.

\noindent
** Requirements **

The name of the step should be one concise noun-phrase.

The abstraction level of the step should be appropriate for high-level planning and communication with non-technical people.
For the parentIds, provide the ids of the steps that this step directly depends on in terms of input-output data.
The alternatives should have varying complexity, coherence with previous steps, and importance.

The whole pipeline should emphasize concise and clear steps, with as few steps as possible and no more than 5 steps.

DO NOT output steps like data collection, implementation, validation or any steps related to communication such as visualization or reporting.

Reply with this JSON format. Do not wrap the code in JSON markers. Do not include any comments.
\begin{lstlisting}
{
  "steps": [
    {
      "label": (string) or "END"
      "description": (string),
      "explanation": (string, explain why this step is needed),
      "parentIds": (string[], ids of the steps that this step **directly** depends on)
    },
    ... ({n} different next steps)
  ]
}
\end{lstlisting}
\end{promptbox}

(2). Convert one semantic task into one or multiple primitive tasks.
\begin{promptbox}
\noindent
** Context **

You are a Natural Language Processing assistant. You are given a list of primitive NLP tasks that could be used.
Here is the list of primitive NLP tasks:

\verb|{primitive_task_defs}|

\noindent
** Task **

The user will describe a series of real-world tasks (semantic tasks). Your job is to convert each semantic task into one or more primitive NLP tasks necessary to accomplish the goal. You will process one semantic task at a time, considering previously generated primitive tasks AND the description of the next semantic task to avoid overlap.

\noindent
** Requirements **

1. The ids of each formulated NLP task must be unique. The "label" field in your output MUST ONLY use one of these exact labels from the primitive task list.
\verb|{examples for ids and labels}|

2. A single semantic task often requires MULTIPLE primitive tasks chained together. STRICTLY enforce input/output compatibility between primitive tasks:

\verb|{requirement details}|

3. Correctly handle dependencies WITHIN the current step.

\verb|{requirement details}|

4. MAXIMIZE REUSE of existing primitive tasks from PREVIOUS steps.

\verb|{requirement details}|

5. DO NOT GENERATE PRIMITIVE TASKS FOR FUTURE SEMANTIC TASKS.

\verb|{requirement details}|

\noindent
** Examples of Common Task Chains **

\verb|{examples}|

Reply with the following JSON format:

\begin{lstlisting}
{ 
  "primitive_tasks": [
    {
      "solves": (string) id,
      "label": (string) (MUST be one of {supported_labels}),
      "id": (str) (a unique id for the task),
      "description": (string),
      "explanation": (string),
      "depend_on": (str[]),
    },
    ...
  ],
  "validation_check": "{validation_check}"
}
\end{lstlisting}
\end{promptbox}

(3). Generate structured prompt templates for prompt tools in the executor.
\begin{promptbox}
\noindent
** Context **

You are an expert in writing prompts for Large Language Models, especially for generating prompts that analyze a given piece of text.

\noindent
** Task **

The user will describe the task and provide a piece of text. You need to generate a prompt that can be used to analyze the text for the given task.

\noindent
** Requirements **

First, decide what data in each document is needed to complete the task. Here are the data already exist on each document, with type and schema: 

\verb|{existing_keys}|

Then, generate a prompt that instructs an LLM to analyze the text using the data in each document for the user's task. The prompt should be a JSON object with these three sections:

1. Context: Give instructions on what the user is trying to do.

2. Task: Give instructions on how to analyze the text.

3. Requirements: Provide any specific requirements or constraints for the prompt.

4. \verb|JSON_format|: A JSON object with one key, the key name should be suitable to store the result of the prompt, and value should be a valid JSON format for representing the output. The key name of \verb|JSON_format| should be DIFFERENT from any following keys: [ \verb|{all_keys_str}| ]

5. \verb|output_schema|: The "\verb|output_schema|" key should provide a detailed description of the output structure defined for the key in \verb|JSON_format|, using the clearer schema notation.

\verb|{output_schema examples}|

Reply with this JSON format:

\begin{lstlisting}
{
    "prompt": {
        "Context": str,
        "Task": str,
        "Requirements": str
        "JSON_format": str
    },
    "output_schema": str
}
\end{lstlisting}
\end{promptbox}

(4). Identify and generate input keys and schema required to perform a primitive task.
\begin{promptbox}
\noindent
** Context **

You are an expert in data schema design and text analytics.

\noindent
** Task **

The user will describe a task for you, and what data is available in the dataset. Your task is to pick the keys that are required from the dataset to complete the task, along with their detailed schema definition. The keys will be grouped by their state, and you must select keys only from ONE state.

\noindent
** Requirements **

You MUST only select keys inside the \verb|<state></state>| block from the \verb|<existing_keys></existing_keys>| block. No additional keys outside this list can be used. All selected keys should belong to the same state.

\verb|{single_key_instruction}|

For each key, provide a detailed schema definition that describes the exact structure.

\verb|{schema_examples}|

Reply with this JSON format:

\begin{lstlisting}
{
    "required_keys": [
        {
            "key": str,
            "schema": str or object
        }
    ]
}
\end{lstlisting}
\end{promptbox}

(5). Generate prompt templates for result evaluators to evaluate task execution results.
\begin{promptbox}
\noindent
** Context **

You are an expert in generating LLM judges. An LLM judge is an agent that can generate a score using some criteria.

\noindent
** Task **

The user will describe a text analysis task that he/she has done on a list of documents, and how he/she wants the llm judge to evaluate the task result of each document. Your task is to generate a specification of the LLM judge.

The LLM judge should do the following: First, identify what the user wants to evaluate on each document. Then, for each document, output a categorical score using the user-specified criteria.

\noindent
** Requirements **

In the LLM judge's prompt template, specify that the LLM judge must generate only **ONE** score for each document. Each score in "Possible Scores" should be a single word or at most a short noun-phrase.

Reply the evaluator specification with this JSON format.

\begin{lstlisting}
{
    "evaluator_specification": {
        "name": str,
        "definition": str,
        "prompt_template": {
                "Context": str,
                "Task": str,
                "Possible Scores": list[str]
        }
    }
}
\end{lstlisting}
\end{promptbox}

(6). Generate recommendations on criteria for evaluating the task execution results.
\begin{promptbox}
\noindent
** Context **

You are an expert in generating evaluation criteria for text analysis tasks.

\noindent
** Task **

The user will describe one text analysis task that he/she has done on a list of documents along with the overarching goal.

Your task is to recommend up to three criteria that can be used to evaluate the task result of each document.

\noindent
** Requirements **

Reply with the following JSON format:

\begin{lstlisting}
{
    "evaluator_descriptions": [{
        "name": str,
        "description": str,
    }]
}
\end{lstlisting}
\end{promptbox}

(7). Generate execution plan and python code for the data transformation tool.
\begin{promptbox}
\noindent
** Context **
 
You are an expert in creating data transformation plans for document processing. You will create a plan for transforming input data from one schema to another based on the task description by writing executable Python code.

\noindent
** Task **

The user will provide a description of a data transformation task along with input data (typically a list of objects) and the schema of each object in the list. Your job is to create a detailed transformation plan based on this information. The transformation plan must include:

- Operation Type: Specify the type of operation (e.g., "transform") based on the task. Default to "transform" if not otherwise specified.

- Python Code: Provide the exact, executable Python code to perform the transformation. Ensure the code is concise, well-commented, and handles the input data correctly according to the described task.

\verb|{python_code_requirements}|

- Output Schema: Define the schema of the transformed data (e.g., the structure of each object in the resulting list) after applying the transformation.

\verb|{output_schema_requirements}|

Common examples and errors:

\verb|{common_examples}|

\verb|{common_errors_to_avoid}|

Reply with this JSON format:

\begin{lstlisting}
{
    "operation": "transform",
    "parameters": {
        "transform_code": str
    },
    "output_schema": str
}
\end{lstlisting}
\end{promptbox}

(8). Generate execution parameters for the clustering tool.
\begin{promptbox}
\noindent
 ** Context **
 
You are an expert in creating clustering plans for document analysis.
You will create a configuration for clustering documents based on the task description.

\noindent
** Task **

The user will describe a clustering task. You need to create a clustering configuration that includes the algorithm to use and the parameters for that algorithm. You need to carefully select the parameters for clustering based on the input data schema provided by the user.

Also, you also need to determine the key for "\verb|output_schema|" to define the clustering output labels.

\verb|{output_schema_requirements}|

\noindent
** Note **

The input data needs to contain embeddings (vector representations) for effective clustering.

\noindent
** IMPORTANT: Available Clustering Algorithms **

Choose ONE of these algorithms based on the task requirements:

1. kmeans - K-means clustering (requires number of clusters)

\verb|{k_means_description_and_parameters}|

2. dbscan - Density-Based Spatial Clustering (doesn't require number of clusters)

\verb|{dbscan_description_and_parameters}|

3. agglomerative - Hierarchical clustering

\verb|{agglomerative_description_and_parameters}|

4. \verb|gaussian_mixture| - Gaussian Mixture Model

\verb|{gaussian_mixture_description_and_parameters}|

5. hdbscan - Hierarchical DBSCAN

\verb|{hdbscan_description_and_parameters}|

6. bertopic - BERTopic (uses transformer models with UMAP+HDBSCAN)

\verb|{bertopic_description_and_parameters}|

Reply with this JSON format:

\begin{lstlisting}
{
    "algorithm": "kmeans" | "dbscan" | "agglomerative" | "gaussian_mixture" | "hdbscan" | "bertopic",
    "parameters": {
        // Parameters specific to the chosen algorithm
    },
    "output_schema": "{ '<your generated output key>': 'int'}"
}
\end{lstlisting}
\end{promptbox}

(9). Generate execution parameters for the dimensionality reduction tool.
\begin{promptbox}
\noindent
** Context **

You are an expert in creating dimensionality reduction plans for data visualization and analysis. You will create a configuration for reducing the dimensions of data based on the task description.

\noindent
** Task **

The user will describe a dimensionality reduction task. You need to create a configuration that includes the algorithm to use and the parameters for that algorithm. You need to carefully select the parameters for dimensionality reduction based on the input data schema provided by the user.

Also, you also need to determine the key for \verb|"output_schema"| to define the dimensionality reduction output.

\verb|{output_schema_requirements}|

\noindent
** IMPORTANT: Available Dimensionality Reduction Algorithms **

Choose ONE of these algorithms based on the task requirements:

1. pca - Principal Component Analysis

\verb|{pca_description_and_parameters}|

2. tsne - t-distributed Stochastic Neighbor Embedding

\verb|{tsne_description_and_parameters}|

3. umap - Uniform Manifold Approximation and Projection

\verb|{umap_description_and_parameters}|

You MUST respond with a JSON object in this exact format:

\begin{lstlisting}
{
    "algorithm": "pca" | "tsne" | "umap",
    "parameters": {
        "n_components": int,  // Number of dimensions to reduce to (typically 2 or 3 for visualization)
        // Additional algorithm-specific parameters
    },
    "output_schema": "{ '<your generated output key>': 'list[float]'}"
}
\end{lstlisting}
\end{promptbox}

(10). Generate execution parameters for the embedding tool.
\begin{promptbox}
\noindent
** Context **

You are an expert in creating embedding plans for text data. You will create a configuration for generating vector embeddings based on the task description.

\noindent
** Task **

The user will describe an embedding task. You need to create a configuration that includes the embedding model to use and the parameters for that model. You need to carefully select the parameters for embedding based on the input data schema provided by the user.

Also, you also need to determine the key for "\verb|output_schema|" to define the embedding output. 

\verb|{output_schema_requirements}|

\noindent
** IMPORTANT: Available Embedding Providers and Models **

Choose a provider and model combination based on the task requirements:

1. openai - OpenAI's embedding models

\verb|{openai_embedding_description_and_parameters}|

2. \verb|sentence_transformers| - Local embedding models using Sentence Transformers

\verb|{sentence_transformers_description_and_parameters}|

You MUST respond with a JSON object in this exact format:

\begin{lstlisting}
{
    "provider": "openai" | "sentence_transformers",
    "parameters": {
        "model": str,  // Embedding model to use
        // Any additional parameters specific to the provider
    },
    "output_schema": "{ '<your generated output key>': 'list[float]'}"
}
\end{lstlisting}
\end{promptbox}

(11). Generate execution parameters for the segmentation tool.
\begin{promptbox}
\noindent
** Context **

You are an expert in text segmentation for document processing. You will create a configuration for segmenting text documents based on the task description.

\noindent
** Task **

The user will describe a segmentation task. You need to create a configuration that includes the segmentation strategy and the parameters for that strategy.

You need to carefully select the parameters for segmentation based on the input data schema provided by the user.

Also, you also need to determine the key for "\verb|output_schema|" to define the segmentation output.

\verb|{output_schema_requirements}|

\noindent
** IMPORTANT: Available Segmentation Strategies **

Choose ONE of these strategies based on the task requirements:

1. paragraph - Split text by paragraphs

\verb|{paragraph_segmentation_description_and_parameters}|

2. sentence - Split text by sentences

\verb|{sentence_segmentation_description_and_parameters}|

3. \verb|fixed_length| - Split text into chunks of specified length

\verb|{fixed_length_segmentation_description_and_parameters}|

4. semantic - Split text based on semantic meaning

\verb|{semantic_segmentation_description_and_parameters}|

You MUST respond with a JSON object in this exact format:

\begin{lstlisting}
{
    "required_keys": [
        {
            "key": str,
            "schema": str or object
        }
    ]
}
\end{lstlisting}
\end{promptbox}

(12). Evaluate if a task is simple or complex based on the provided definition.
\begin{promptbox}
**Context**

You are a task complexity evaluator. The user will provide some text that describes a task.

\noindent
**Task**

Your job is to decide if the task is complex (hard) or NOT complex (easy) based on the following definition:

\verb|{definition}|

If the task is complex, respond with:

"Yes"

If the task is NOT complex, respond with:

"No"

You must output your reasoning in a <REASONING>...</REASONING> block, then provide your final decision in a <RESULT>...</RESULT> block. The <RESULT> block must contain EXACTLY "Yes" or "No" (nothing else).

Example format:

<REASONING>This is my reasoning about complexity.</REASONING>

<RESULT>Yes/No</RESULT>
\end{promptbox}

(13). Evaluate whether a task logically follows from a parent task based on the provided definition of coherence.
\begin{promptbox}
**Context**

You are a coherence evaluator. You will be given two parts of a task sequence: a parent part and a child part.

\noindent
**Task**

Evaluate whether the child part logically or thematically follows from the parent part according to the following definition of coherence:
  
\verb|{definition}|

If the tasks are coherent, respond with:

"Yes"

If the tasks are NOT coherent, respond with:

"No"

You must output your reasoning in a <REASONING>...</REASONING> block, then provide your final decision in a <RESULT>...</RESULT> block. The <RESULT> block must contain EXACTLY "Yes" or "No" (nothing else).

Example format:

<REASONING>This is my reasoning about coherence.</REASONING>
  
<RESULT>Yes/No</RESULT>

\end{promptbox}

(14). Evaluate whether a task is important to achieve the goal based on the provided definition.
\begin{promptbox}
***Context**

You are an importance evaluator. You will be given a final task goal and a subtask description.

\noindent
**Task**

Evaluate whether the subtask is important using the following definition:

\verb|{definition}|

If the subtask is important, respond with:

"Yes"

If the subtask is NOT important, respond with:

"No"

You must output your reasoning in a <REASONING>...</REASONING> block, then provide your final decision in a <RESULT>...</RESULT> block. The <RESULT> block must contain EXACTLY "Yes" or "No" (nothing else).

Example format:

<REASONING>This is my reasoning about importance.</REASONING>

<RESULT>Yes/No</RESULT>
  
\end{promptbox}

(15). Generate a complexity evaluation combining user-provided input with a few-shot reasoning.
\begin{promptbox}
**Context**

You are an expert in evaluating task complexity. 

\noindent
**Task**

The user will provide a task description, and your role is to analyze its difficulty based on the following definition of complexity:

\verb|{definition}|

Think about what makes this task complex or simple. Consider aspects like how many steps it involves, whether it requires specialized knowledge, if there are any hidden challenges, or how much effort it demands. 

If a task is complex, it's usually because it takes more time, needs deeper problem-solving, or has multiple moving parts. If it‘s simple, it might be straightforward, require little expertise, or have a clear path to completion.

Don't just repeat the definition—apply it to the task at hand. Explain your reasoning naturally, as if you were discussing it with someone who asked, "Why is this task complex (or simple)?" Keep it clear and insightful.

\end{promptbox}

(16). Generate a coherence evaluation combining user-provided input with a few-shot reasoning.
\begin{promptbox}
**Context**

You are an expert in evaluating whether two sequential tasks are coherent. 

\noindent
**Task**

The user will provide a parent task and a child task, and your role is to analyze whether the child task logically or thematically follows from the parent task, based on the following definition of coherence:

\verb|{definition}|

When considering coherence, think about whether the transition from one task to the next feels natural. 

Does the child task build upon the parent task in a meaningful way? Are they part of the same larger goal, or does the shift feel abrupt? 

Some tasks are naturally connected, while others might be loosely related or entirely disconnected.

Rather than listing points mechanically, explain your reasoning as if you were talking to someone who asked, “Do these two tasks make sense together?” Offer a clear, thoughtful explanation.

\end{promptbox}

(17).  Generate an importance evaluation combining user-provided input with a few-shot reasoning.
\begin{promptbox}
**Context**

You are an expert in evaluating task importance. The user will provide a final task goal and a subtask description. 

\noindent
**Task**

Your role is to analyze whether the subtask is essential for achieving the final goal based on the following definition of importance:

\verb|{definition}|

When thinking about importance, ask yourself: Does this subtask play a key role in reaching the final goal? 

Is it something that must be done, or is it more of an optional step? Some subtasks are crucial because they provide necessary information, resources, or foundations, while others might be useful but not strictly necessary.

Instead of following a rigid structure, explain your reasoning in a way that feels natural—almost like you're answering someone who asked, "Do you think this step really matters?" Keep it conversational, clear, and insightful.

\end{promptbox}

(18). Summarize multiple reasoning explanations for an evaluation into an overview.
\begin{promptbox}
**Context**

You are a reasoning summarizer. You will be given multiple reasoning explanations about the same evaluation.

\noindent
**Task**

Your task is to synthesize these explanations into a single, coherent summary that captures the key points and rationale.
  
Keep the summary concise but ensure it maintains the core reasoning and important details from the input explanations.

If there are conflicting viewpoints, include both perspectives in your summary.
  
Output only the summarized reasoning, with no additional formatting or meta-commentary.

\end{promptbox}

(19). Generate topic labels summarizing a cluster of texts for radial chart.
\begin{promptbox}
You are a topic assignment system. The user will provide you with a bunch of texts. You need to assign one topic to summarize all of them. The topic should be a simple noun-phrase. Only one topic should be generated. Reply with a single noun-phrase as the topic.
\end{promptbox}

(20). Compare the pros and cons of two pipelines (used in the decomposer evaluation experiment) with OpenAI's o3-mini reasoning model.
\begin{promptbox}
I have a text analytics goal: \verb|{goal}|.
            I have come up with two solutions: 
            Solution 1: \verb|{pipeline_1}|.
            Solution 2: \verb|{pipeline_1}|.
            What do you think about my solutions? 
            Organize the answers with explicit criteria.
            For each criteria, give me the pros and cons of each solution.
            After the whole analysis, pick one solution that you think is the better one.
\end{promptbox}

\end{multicols}

\newpage

\begin{figure*}[t]
    \centering
    \includegraphics[width=\textwidth]{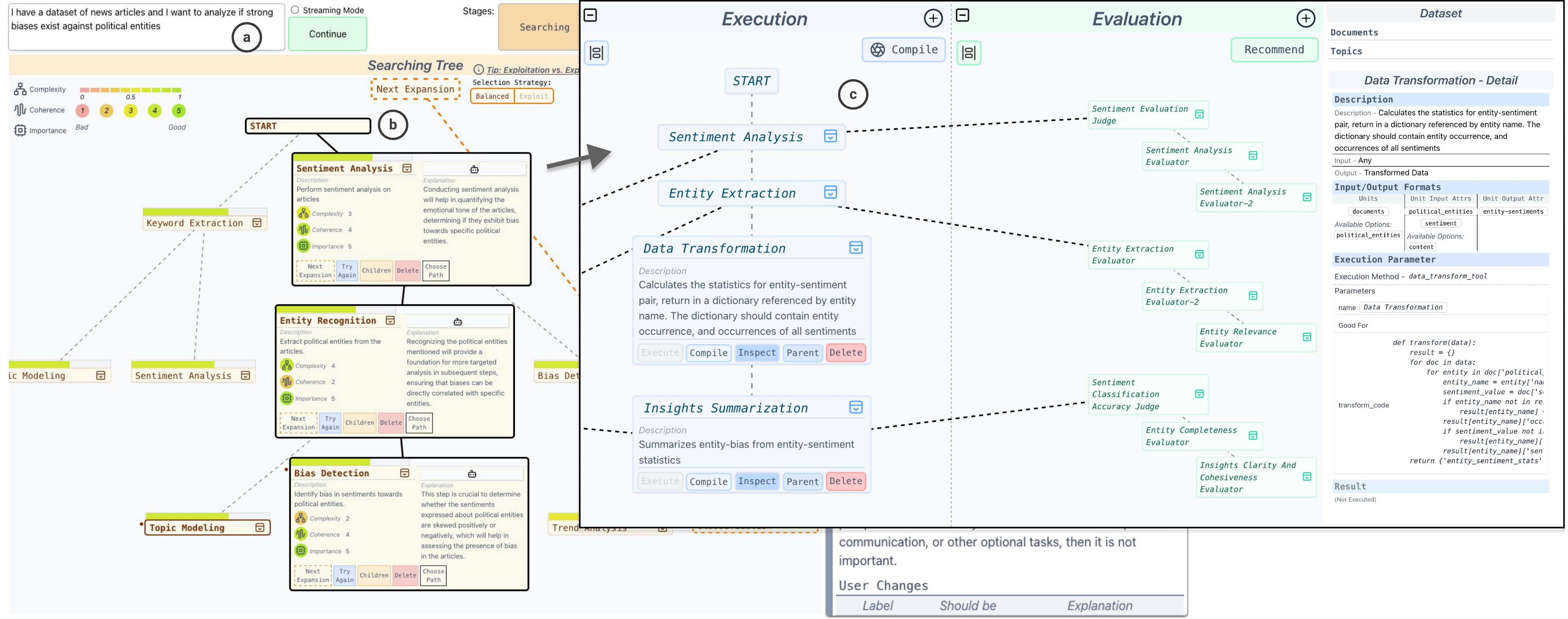}
    \caption{Screenshots for usage scenario: political entity bias analysis. (a) the goal input box where the user enters the goal, (b) the search tree generated by the system and the selected plan, (c) the execution page specifying the executable pipieline and relevant paramters.}
    \label{fig: use_case_page}
\end{figure*}

\section{Usage scenario: Political Entity Bias Analysis}

\begin{wrapfigure}{r}{0.45\textwidth}
    \centering
    \includegraphics[width=\linewidth]{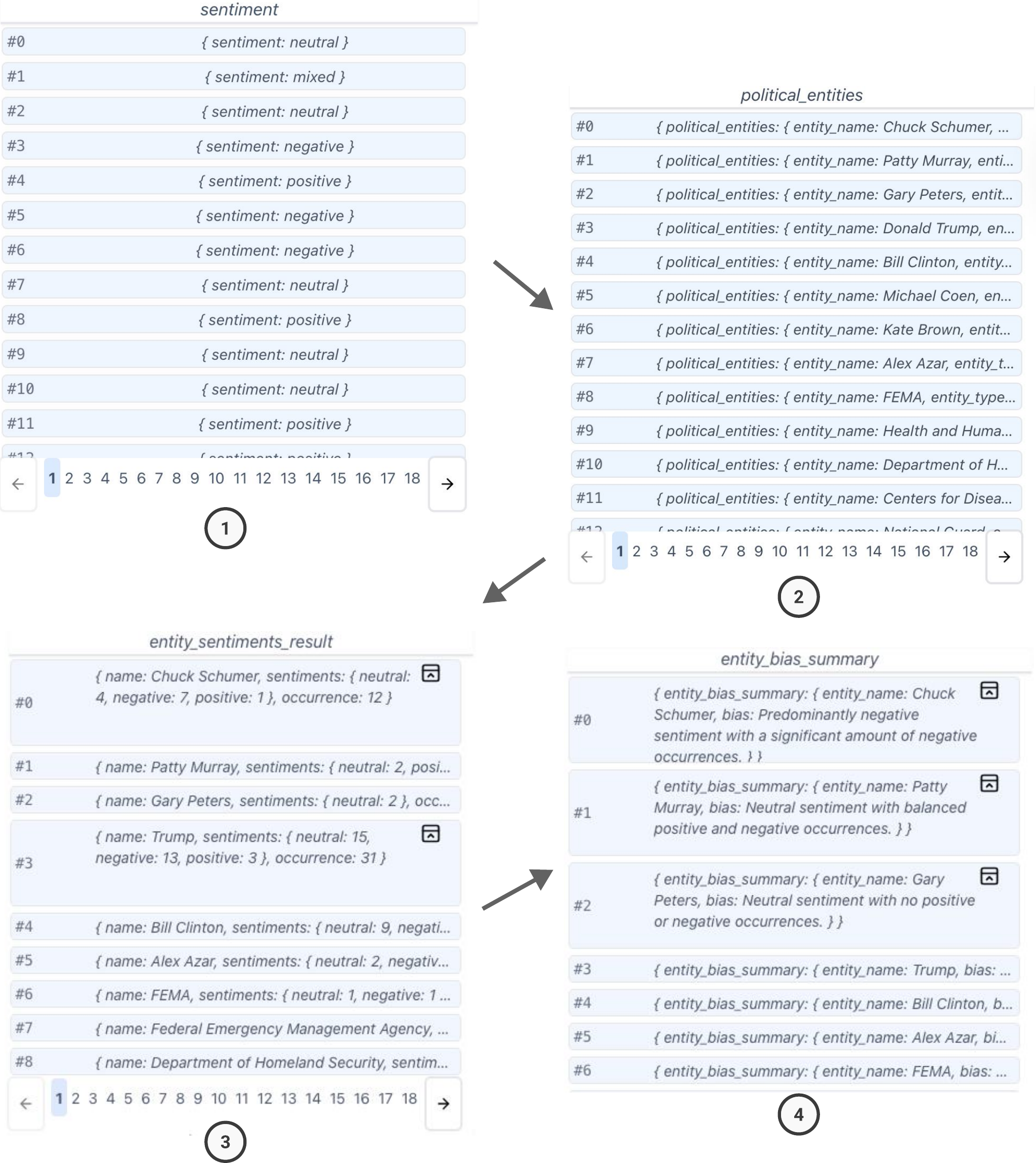}
    \caption{Screenshots of the intermediate data structure changes after each step. (1) After sentiment analysis, each document is assigned a sentiment label, (2) After entity extraction, each document is assigned a list of entities, (3) After data transformation, entity and sentiment occurrences are aggregated from all documents, (4) After insight summarization, each entity is analyzed with potential biases from the sentiments.}
    \label{fig: use_case_result}
\end{wrapfigure}

We demonstrate how \system \ can be used to analyze biases on political entities on a news article dataset.

Alice, a data analyst given the task, enters the goal directly into the goal input box (\autoref{fig: use_case_page}-a). The system generates a few alternatives of plans, and Alice chooses one of the path consisting of \textit{Sentiment Analysis}, \textit{Entity Recognition}, and \textit{Bias Detection}, as it provides the most reasonable and interpretable results (\autoref{fig: use_case_page}-b). Alice then moves on the the execution stage.

At the execution stage (\autoref{fig: use_case_page}-c), Alice converts the plan into an executable pipeline by clicking the ``Convert'' button. The system generates a pipeline with four steps: \textit{Sentiment Analysis}, \textit{Entity Extraction}, \textit{Data Transformation}, and \textit{Insights Summarization}. Each step comes with a clear description of its purpose and necessary paramters.

Upon executing the tasks step-by-step, Alice can see how the results change ((\autoref{fig: use_case_result})). In the first two steps, each document is assigned with a label ``sentiment'' and a value of ``$<$negative, neutral, positive$>$'', and corresponding entities. Alice can see that the extracted entities are indeed all politics-related. In the third step, entities and sentiments from each document are aggregated for entity-sentiment occurences. Finally, the fourth step summarizes bias-related insights from the entity-sentiment occurrences. 

With the help of \system, Alice doesn't need to write codes to implement sentiment analysis, entity extraction, and data transformation, and can easily generate and execute the pipeline with simple user interactions.

\end{document}